\documentclass{article}

\usepackage{report} 

\usepackage[utf8]{inputenc} % 允许 utf-8 输入
\usepackage[T1]{fontenc}    % 使用 8-bit T1 字体
\usepackage[table,dvipsnames,svgnames]{xcolor} 
\usepackage{graphicx}
\usepackage{transparent} % 透明度支持

\usepackage{amsmath}
\usepackage{amssymb}
\usepackage{amsfonts}
\usepackage{nicefrac}       % 紧凑的 1/2 符号
\usepackage{microtype}      % 微排版优化
\usepackage{soul}           % 字符高亮/划线

\usepackage{booktabs}       % 专业表格线
\usepackage{multirow}
\usepackage{array}
\usepackage{longtable}
\usepackage{enumitem}       % 列表定制
\usepackage{fancyhdr}       % 页眉页脚
\usepackage{wrapfig}
\usepackage{float}
\usepackage{subcaption}
\usepackage{placeins}       % 控制浮动体位置
\usepackage{ragged2e}       % 文本对齐

\usepackage{pifont}
\usepackage{utfsym}
\usepackage{fontawesome5}   % 建议使用 v5，移除旧版 fontawesome 以免冲突
\usepackage{CJKutf8}        % 中文支持

\usepackage{listings}
\usepackage[most]{tcolorbox} % 'most' 加载了 skins, breakable 等常用库
\tcbuselibrary{breakable}

\usepackage{lipsum}         % 生成示例文本
\usepackage{url}            % 简单的 URL 排版

\usepackage{hyperref}

\definecolor{darkmagenta}{rgb}{0.56, 0.0, 1.0}
\definecolor{softyellow}{rgb}{1.0, 0.92, 0.3}
\definecolor{LightAquamarine}{rgb}{0.75, 1.0, 0.8}
\definecolor{FireBrick}{RGB}{178,34,34}
\definecolor{MediumPurple}{RGB}{147,112,219}
\definecolor{uclablue}{rgb}{0.15, 0.45, 0.68}
\definecolor{njuPurple}{RGB}{220,205,230}     
\definecolor{njuPurpleLight}{RGB}{250,245,252}

\hypersetup{
    breaklinks,
    colorlinks=true,
    citecolor={darkmagenta},
    linkcolor={uclablue},
    urlcolor={uclablue}
}

\newtcolorbox{abstractbox}{
    colback=njuPurpleLight,
    colframe=njuPurple,
    boxrule=1pt,
    arc=4mm,
    left=8pt, right=8pt, top=8pt, bottom=8pt,
    skin=enhanced, % 需要 skins 库 (由 [most] 加载)
    opacityback=0.95
}

\newtcolorbox{promptbox}[1]{
    breakable,
    colback=gray!15!white,
    colframe=gray!75!black,
    fonttitle=\bfseries,
    title=#1,
    bottom=1cm,
    arc=2mm,
    boxrule=0.5pt,
    left=2mm, right=2mm, top=2mm
}

\tcbset{
    mainbox/.style={
        colback=gray!10,
        colframe=black,
        boxrule=0.8pt,
        arc=2mm,
        outer arc=2mm,
        left=6pt, right=6pt, top=6pt, bottom=6pt,
        boxsep=4pt,
        breakable
    },
    titlebox/.style={
        colback=black!85,
        colframe=black,
        boxrule=0.8pt,
        arc=1mm,
        outer arc=1mm,
        left=4pt, right=4pt, top=2pt, bottom=2pt,
        fontupper=\bfseries\color{white}
    }
}

\lstdefinestyle{jsonstyle}{
    basicstyle=\ttfamily\small,
    showstringspaces=false,
    breaklines=true,
    breakatwhitespace=true,
    backgroundcolor=\color{black!5},
    numbers=none,
    commentstyle=\color{gray},
    keywordstyle=\color{blue},
    stringstyle=\color{purple!80!black},
}

\DeclareRobustCommand{\titlelogo}{\smash{\raisebox{-0.25\height}{\includegraphics[width=1.5cm]{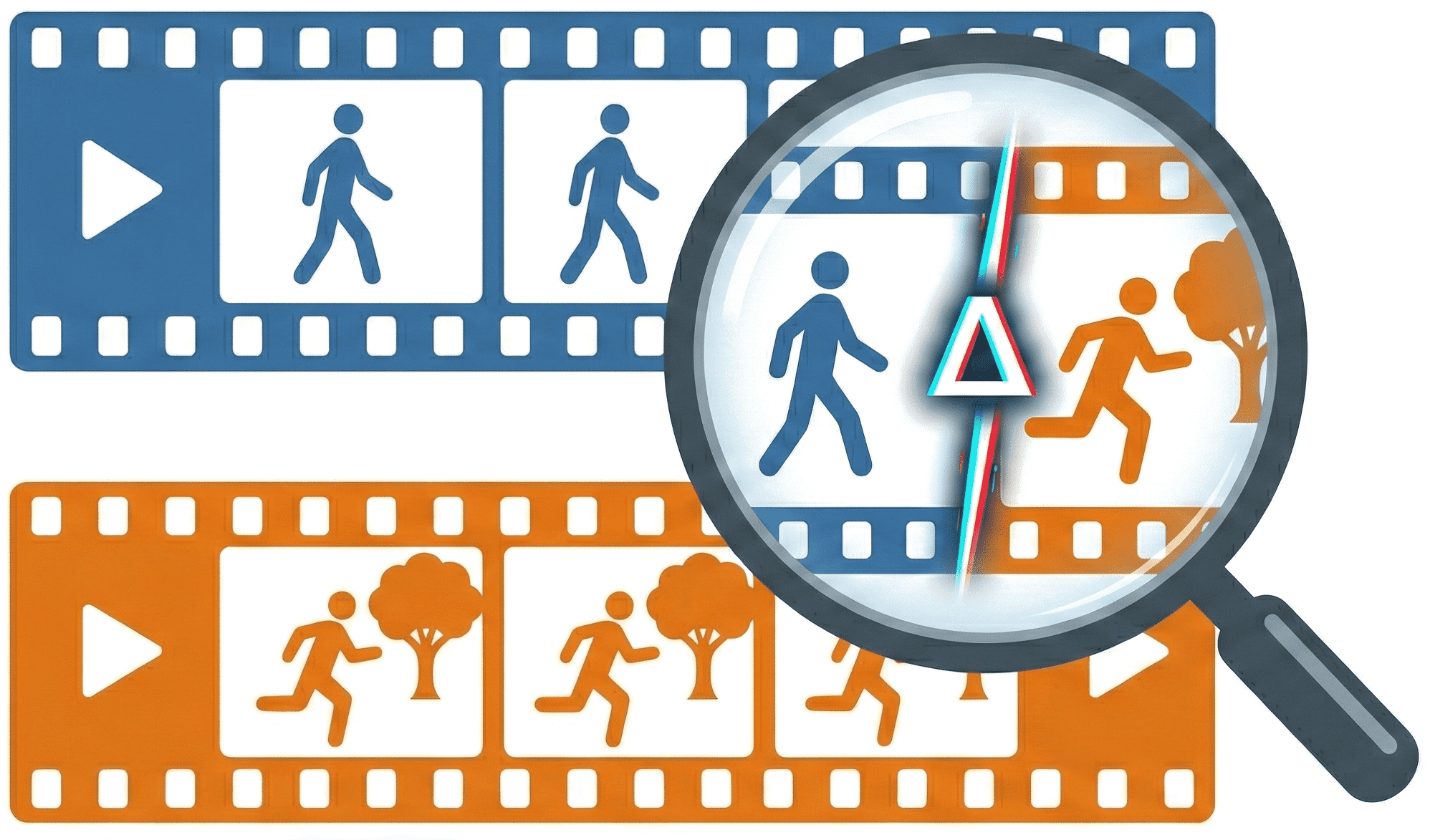}}}}

\title{
\titlelogo
ViDiC: Video Difference Captioning
}

\author{
\textbf{Jiangtao Wu$^{1*}$},
\textbf{Shihao Li$^{1*}$},
\textbf{Zhaozhou Bian$^{1*}$}, \\
\textbf{Jialu Chen$^{2}$},
\textbf{Runzhe Wen$^{1}$},
\textbf{An Ping$^{1}$},
\textbf{Yiwen He$^{1}$},
\textbf{Jiakai Wang$^{2}$}, \\
\textbf{Yuanxing Zhang$^{2, \dagger}$},
\textbf{Jiaheng Liu$^{1,\dagger}$} \\
\vspace{4mm}
{\normalsize  $^1$ NJU-LINK Team, Nanjing University} \quad
{\normalsize  $^2$ Kling Team, Kuaishou Technology} \\
\vspace{2mm}
\texttt{jiangtaowu@smail.nju.edu.cn}
\quad\quad\quad
\texttt{liujiaheng@nju.edu.cn} \\
}

\begin{document}

\maketitle

\let\oldthefootnote\thefootnote
\let\thefootnote\relax\footnotetext{*~Equal Contribution. ~~$^\dagger$~Corresponding Author.}
\let\thefootnote\oldthefootnote

\begin{abstractbox}
\begin{center}
\textbf{\Large Abstract}
\end{center}

Understanding visual differences between dynamic scenes requires the comparative perception of compositional, spatial, and temporal changes—a capability that remains underexplored in existing vision-language systems. While prior work on Image Difference Captioning (IDC) has enabled models to describe semantic changes between static images, these approaches fail to capture motion continuity, event evolution, or video editing consistency over time. We introduce the \textbf{ViDiC (Video Difference Captioning)} task and its corresponding \textbf{ViDiC-1K} dataset, designed to evaluate the ability of Multimodal Large Language Models (MLLMs) to provide fine-grained descriptions of similarities and differences between video pairs. ViDiC-1K comprises 1,000 curated video pairs annotated with 3720 comparative checklist items, covering seven categories: subject, style, background, camera work, motion, position, and playback techniques. To ensure reliable evaluation, we propose a dual-checklist framework that measures the accuracy of similarity and difference separately, based on the LLM-as-a-Judge protocol. Experiments on 17 representative multimodal models reveal a significant performance gap in their comparative description and difference perception abilities. We hope ViDiC-1K can be a challenging benchmark that lays a solid foundation for advancing video understanding, edit awareness, and comparative reasoning in multimodal intelligence. The dataset is now open-source and available at \url{https://huggingface.co/datasets/NJU-LINK/ViDiC-1K}.
  
\end{abstractbox}

\begin{figure*}[h]
    \centering
    \includegraphics[width=\textwidth, keepaspectratio]{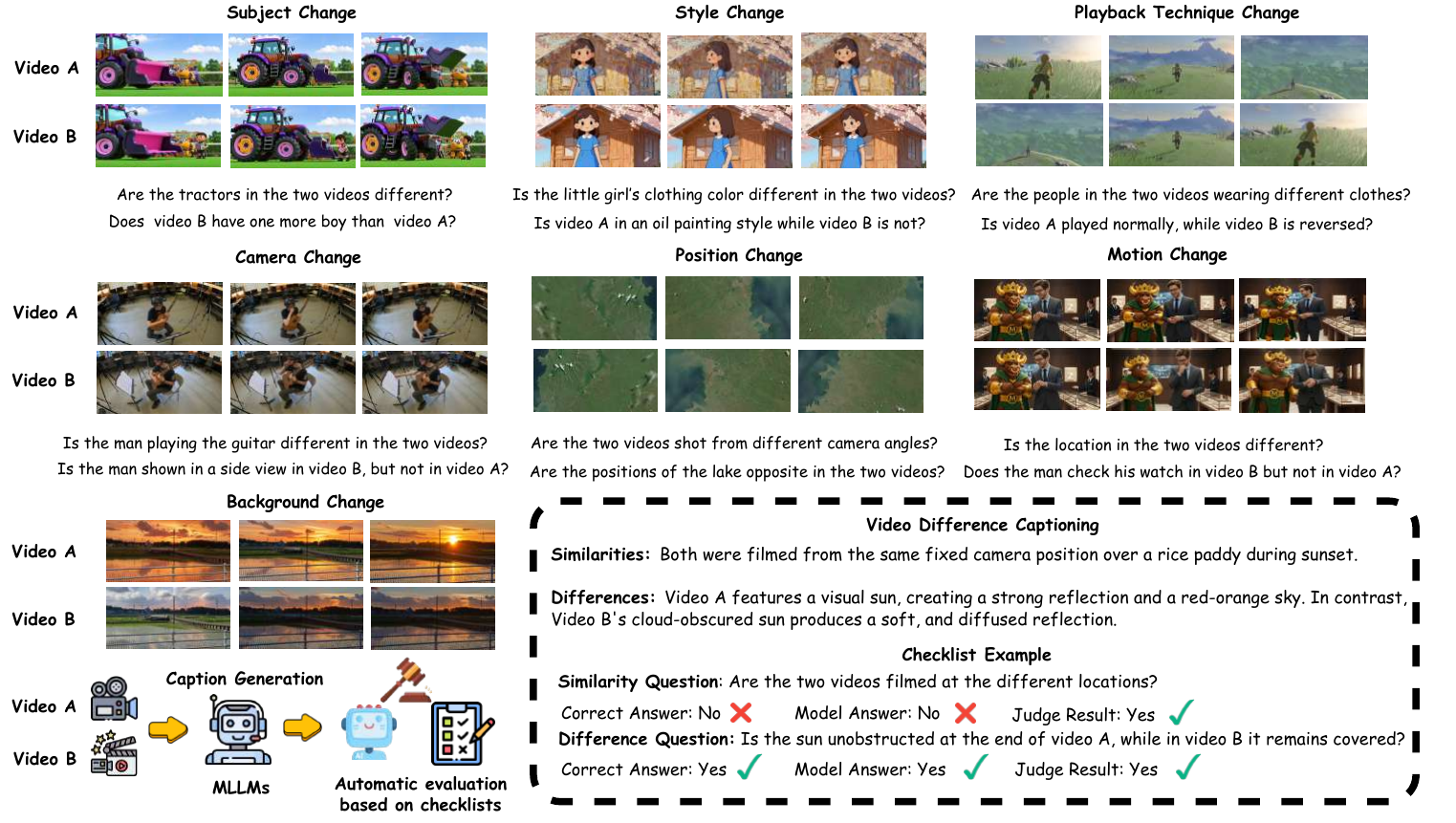}
    \caption{Illustration of the seven categories of video-pair variations in our study: Subject, Style, Playback Technique, Camera Work, Position, Motion, and Background. Using ``Background Change'' as an exemplar, we showcase our Video Difference Captioning task, where a model generates a caption detailing similarities and differences. The caption's accuracy is then assessed against a fine-grained checklist.}
    \label{fig:page}
\end{figure*}

\section{Introduction}
\label{sec:intro}

Understanding and describing differences between visual inputs is a fundamental capability of human perception and a cornerstone of visual reasoning. While recent progress in image difference captioning (IDC)~\citep{jhamtani2018learning, Park2019RobustCC,yao2022idc,di2025difftell,Liu2025OmniDiff} has enabled models to articulate semantic changes between pairs of static images, these methods remain inherently limited: they operate on snapshots, ignoring the temporal evolution and motion cues that define real-world visual experiences. In dynamic scenes, differences are not only found in static frames but also emerge over time — arising from variations in actions, events, camera movements, or stylistic transitions across time. 
To bridge this gap, as shown in Figure~\ref{fig:page}, we introduce \textbf{Video Difference Captioning (ViDiC)}, a new task that extends difference captioning into the video domain. 
Specifically, the ViDiC task requires models to generate natural language descriptions that accurately capture differences in both their static visual content and their temporal dynamics between two video clips while maintaining coherence and factual grounding. 
This formulation moves beyond traditional video similarity~\citep{zeng2019tencent, liberatori2025convis} or video editing evaluation~\citep{tgve2023,argaw2022anatomy,ju2025editverse} tasks, focusing instead on edit understanding rather than edit execution.

However, constructing such a benchmark for the ViDiC task presents several challenges. First, video annotation is costly and ambiguous: differences may arise from subtle temporal cues or stylistic variations not easily expressible in simple labels. Second, existing video editing datasets emphasize task completion metrics (e.g., edit fidelity) \citep{chen2025iveb, sun2024bench, li2025omnivideobenchaudiovisualunderstandingevaluation, pan2025mtvideobenchholisticvideounderstanding, chen2025vidcapbenchcomprehensivebenchmarkvideo}, which fail to capture descriptive capabilities. Finally, scalable benchmarking requires standardized evaluation protocols to ensure consistency across diverse models and data sources.

To address these issues, we introduce a dual-checklist evaluation framework and a high-quality video pair dataset called \textbf{ViDiC-1K}, which is designed explicitly for video difference captioning. Our dataset comprises 1,000 curated video pairs (both real and synthetic), annotated with 3720 fine-grained comparative questions spanning seven semantic dimensions: \textit{Subject, Style, Background, Camera Work, Motion, Position, and Playback Technique.} Each pair is accompanied by both similarity and difference checklists, enabling a detailed, interpretable assessment that transcends single-score metrics. Beyond data design, we propose an LLM-assisted evaluation protocol where a large judge model (GPT-5-Mini) quantifies factual accuracy by comparing generated captions against human-verified ground truths. This scalable, model-agnostic evaluation paradigm ensures reliable comparison across systems without requiring direct visual access during judgment.

% Through extensive experiments on twelve state-of-the-art models — including GPT-5, Gemini-2.5-Pro, InternVL-3.5, and Qwen-3-VL — we demonstrate that ViDiC exposes crucial performance gaps in fine-grained temporal reasoning and edit interpretation. Our analysis further reveals domain-specific weaknesses (e.g., playback manipulation, motion trajectory, and stylistic consistency) that remain unsolved even for leading MLLMs.

In summary, our contributions are threefold:
% The main contributions can be summarized as follows:
\begin{itemize}
 \item We introduce the \textbf{Video Difference Captioning} task by unifying descriptive, comparative, and temporal understanding, which generalizes image-level difference captioning into the temporal domain and establishes a foundation for advancing multimodal models toward more robust and explainable video reasoning.

\item To evaluate the capabilities of existing MLLMs, we first propose the ViDiC-1K benchmark, comprising 1,000 annotated video pairs with structured similarity–difference checklists across seven spatio-temporal dimensions, and introduce a scalable evaluation framework leveraging LLM-as-a-judge for factual, interpretable, and reproducible benchmarking of MLLMs.

\item Through extensive experiments on existing models, we demonstrate that ViDiC exposes crucial performance gaps in fine-grained temporal reasoning and edit interpretation, and reveals domain-specific weaknesses that remain unsolved even for leading MLLMs.

\item To enhance the model’s capability in ViDiC, we propose a training set, a video difference understanding dataset featuring over 60,000 diverse video pairs across various scenarios. By fine-tuning Qwen-2.5-VL-7B-Instruct on this dataset, we achieve significant performance improvements over the baseline. These results validate the effectiveness of our data and establish a strong foundation for future research in video difference captioning.
\end{itemize}
\section{Related Works}
\label{sec:rel}

\paragraph{\textbf{Visual Difference Understanding.}}
Visual difference understanding has evolved from low-level pixel comparisons, exemplified by Change Detection\citep{8451652,Chen2020ASA} using Siamese architectures, to high-level semantic reasoning\citep{jhamtani2018learning,wu2025editreward}. This shift underscores the necessity of interpreting visual semantics in natural language, a domain where the current generation of MLLMs demonstrates remarkable proficiency across a spectrum of tasks, ranging from canonical captioning and VQA~\citep{10.1007/978-3-319-10602-1_48, 9009481, Hudson_2019_CVPR, okvqa, singh2019towards} to holistic reasoning~\citep{liu2024mmbench, li2024seed2plus}. However, despite these advancements, MLLMs typically operate on a single visual input, restricting their ability to perform comparative understanding. This limitation creates a disconnect with real-world applications—such as video editing verification, content forensics, and intelligent surveillance—where distinguishing fine-grained discrepancies between reference and target footage is essential. Consequently, there is an urgent need to bridge the gap between these practical multi-video requirements and the current deficiency in comparative reasoning capabilities~\citep{peng2025mvuevalmultivideounderstandingevaluation}.

\paragraph{\textbf{Image Difference Captioning.}} Image Difference Captioning task focuses on describing semantic changes between two images. Early studies, such as Spot-the-Diff~\citep{jhamtani2018learning} introduced datasets for learning to verbalize visual differences. Recent works attempt on synthetic data generation and preference-based selection~\citep{wu2025editreward, ju2025editverse,dunlap2024describingdifferencesimagesets}.
Despite these advances, IDC methods rely on static image pairs and thus fail to capture temporal dynamics or motion consistency. In contrast, ViDiC-1K extends the task into the temporal domain by introducing a benchmark for video difference captioning, where models must reason over both spatial and temporal variations between two video clips. Compared to IDC, this task requires understanding event evolution and motion patterns over time, providing a more comprehensive evaluation framework for video understanding and video editing models.

\paragraph{\textbf{Video Editing Datasets.}} The scarcity of high-quality training data remains a primary bottleneck for video editing, constrained by the difficulty of maintaining spatio-temporal consistency and semantic alignment. Although recent works have turned to large-scale synthetic curation~\citep{ju2025editverse}, ensuring the quality of these automated instruction-edit pairs is difficult without a reliable verification loop. This necessitates the development of Video Difference Captioning models, which are essential for scaling up data production by automatically verifying edit success or generating precise edit instructions from video pairs. Unlike prior datasets focused on edit fidelity~\citep{tgve2023,argaw2022anatomy}, ViDiC-1K addresses this need by establishing a benchmark specifically for fine-grained video difference understanding.
\section{ViDiC-1K}
\label{sec:ViDiC}

To evaluate video comparison for editing, we introduce the ViDiC-1K benchmark, built on a new framework derived from editing workflows. The framework organizes comparison criteria into seven super-categories (Figure~\ref{fig:page}) and uses a dual-checklist design assessing both similarities and differences for each video pair. This enables a granular evaluation that overcomes the limitations of single, coarse-grained similarity scores.

\subsection{Data Collection}
\label{sec:Data collection}

\subsubsection{Video Collection}
\label{sec:video_collection}

\begin{figure*}[!t]
    \centering

    % (a) 顶部图片
    \subcaptionbox{Statistical distribution across various categories including Subject, Style, Background, Camera, Motion, Position, and Playback Technique.\label{fig:top}}{%
      \includegraphics[width=\textwidth]{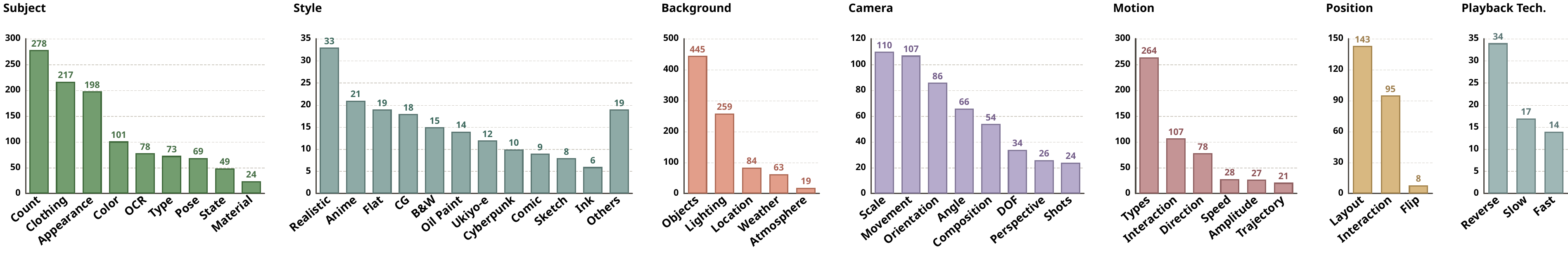}%
    }
  
    \vspace{3mm} % 垂直间距

    % (b) 左列图片
    \subcaptionbox{Hierarchical classification of video content.\label{fig:left}}{%
        \begin{minipage}[b]{0.35\textwidth}
        \includegraphics[width=\linewidth, keepaspectratio]{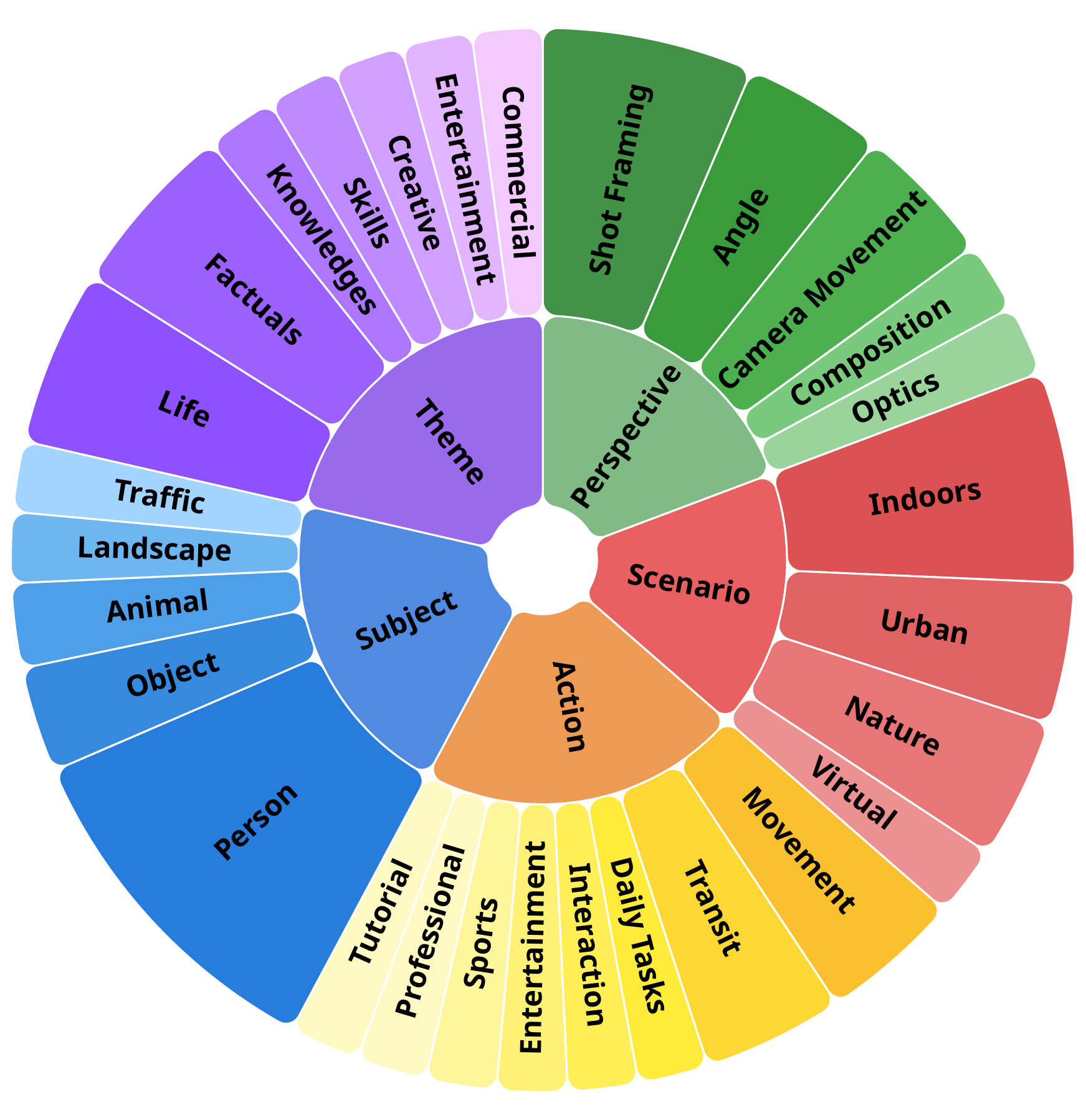}
        \end{minipage}%
    }
    \hfill % 列间距
    % (c, d, e) 中间列，它本身是垂直堆叠的
    \begin{minipage}[b]{0.34\textwidth}
        % (c) 中间-上
        \subcaptionbox{Distribution of the number of checklist items per video pair.\label{fig:middle_top}}{%
            \includegraphics[width=\linewidth]{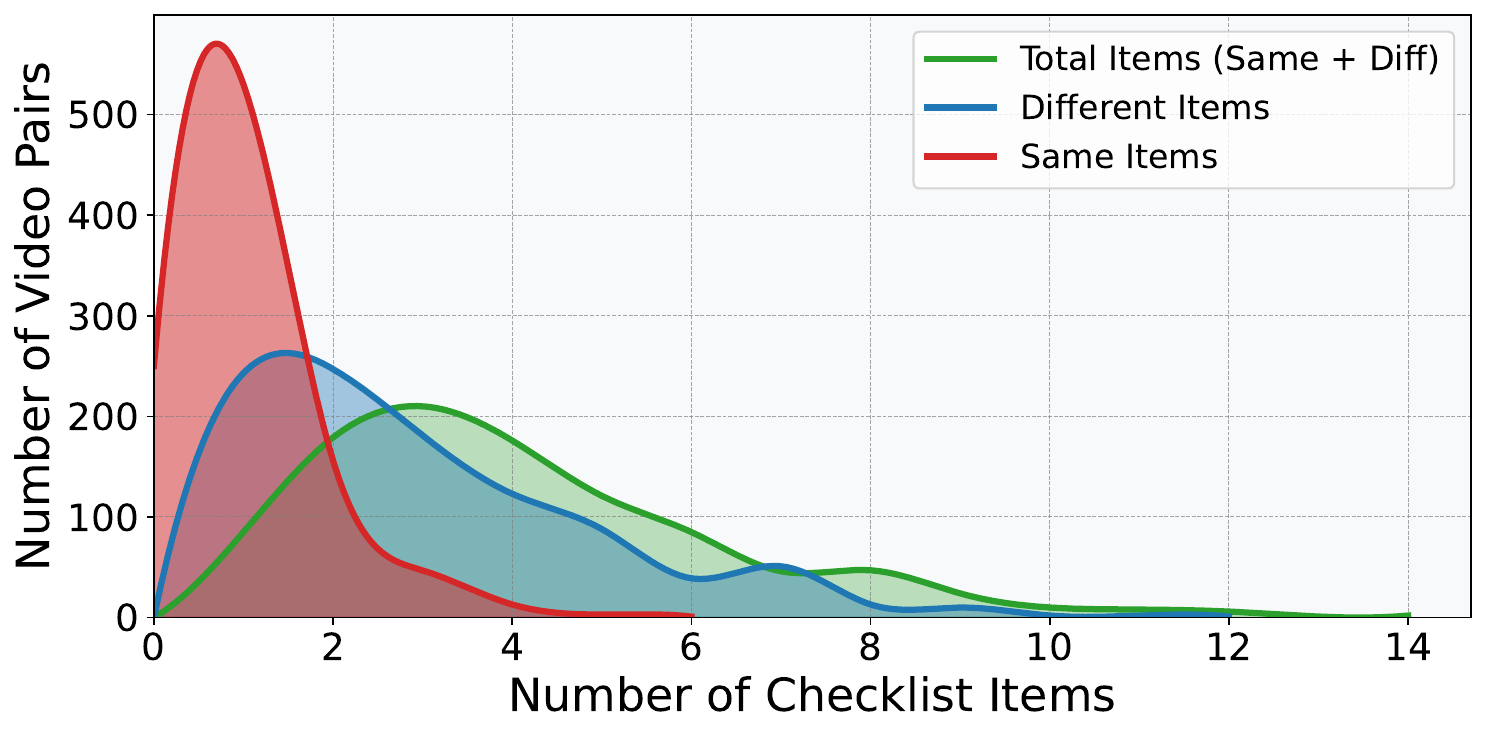}%
        }
        
        \vspace{1.5mm} % 内部垂直间距

        % (d) & (e) 中间-下，并排
        \subcaptionbox{Distribution of video durations.\label{fig:duration}}{%
            \includegraphics[width=0.49\linewidth]{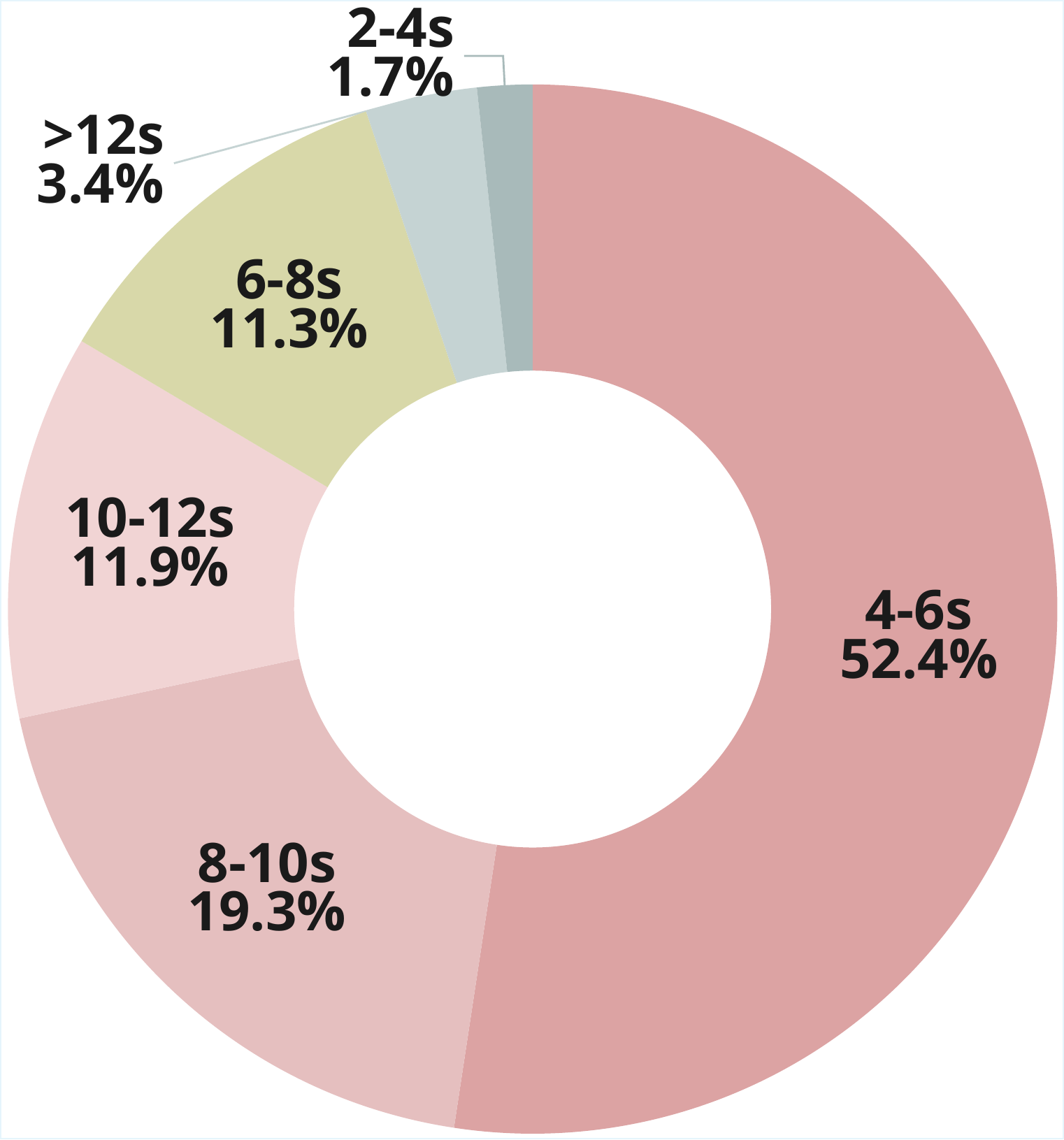}%
        }%
        \hfill
        \subcaptionbox{Distribution of video resolutions.\label{fig:resolution}}{%
            \includegraphics[width=0.49\linewidth]{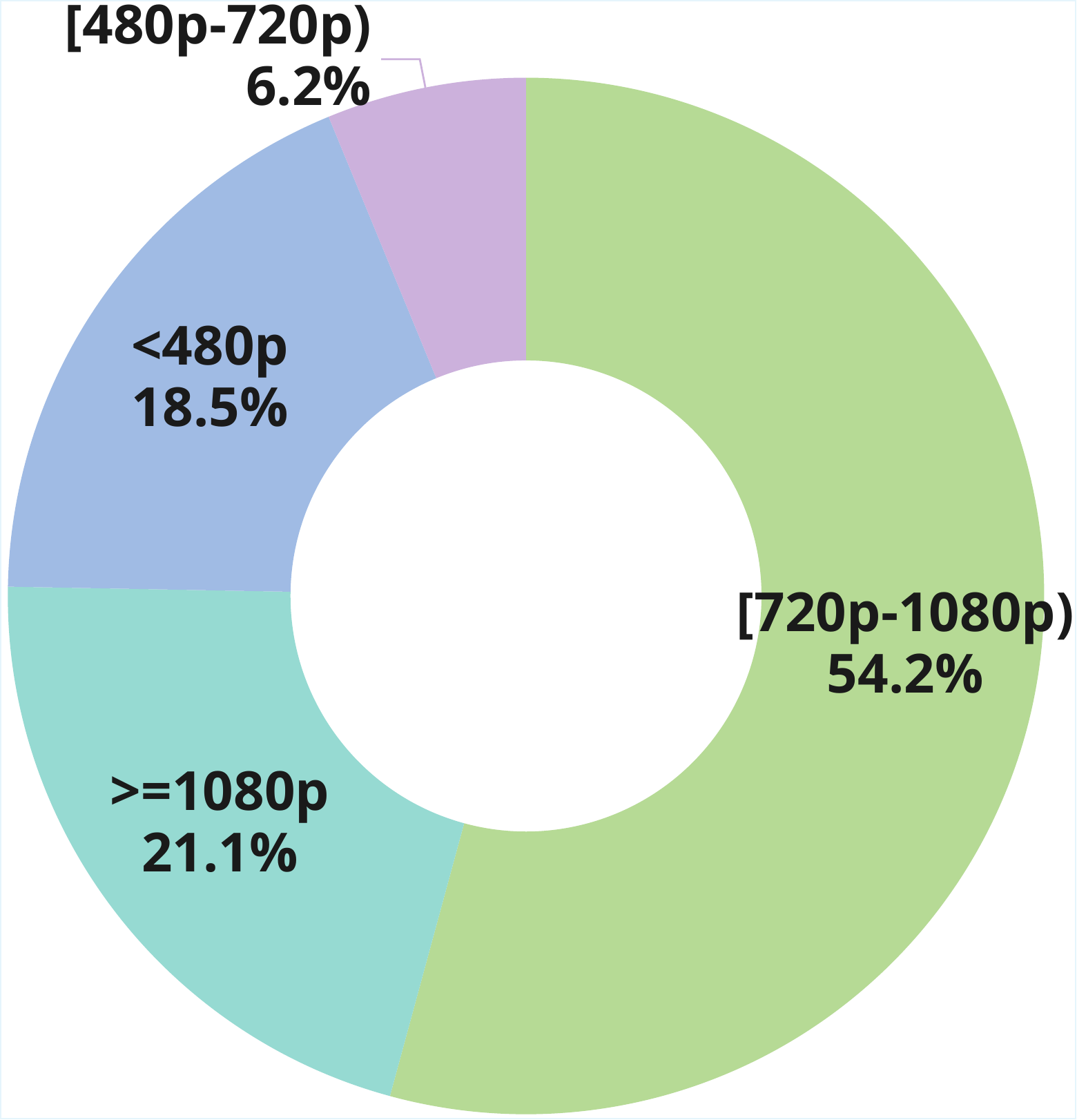}%
        }%
    \end{minipage}
    \hfill % 列间距
    % (f) 右列图片
    \subcaptionbox{Distribution of video source datasets.\label{fig:right}}{%
        % 使用 minipage 来设定宽度，并确保其内部是底部对齐的图像
        \begin{minipage}[b]{0.25\textwidth}
        \includegraphics[width=\linewidth, keepaspectratio]{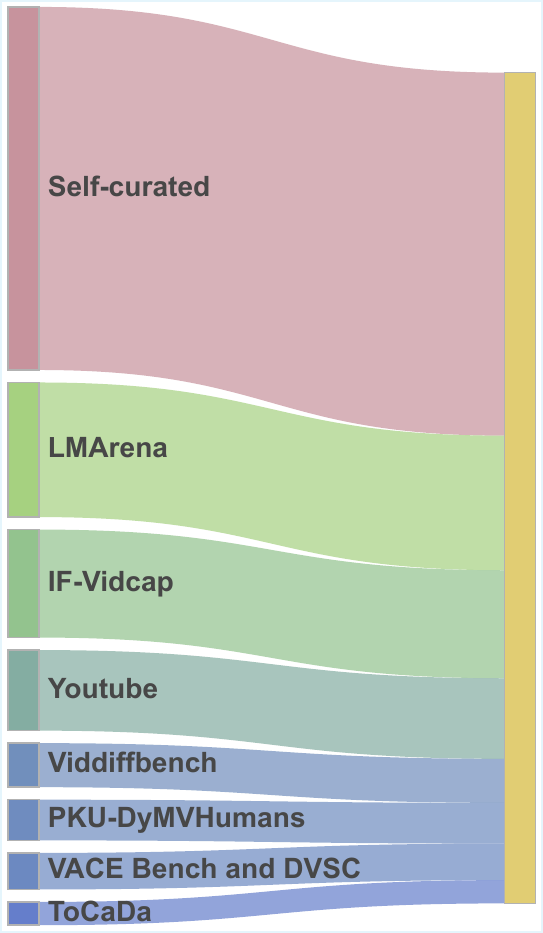}
        \end{minipage}%
    }

    \caption{An overview of the statistical analysis of our dataset across multiple dimensions.}
    \label{fig:full_layout}
\end{figure*}

To establish a benchmark with broad coverage, we constructed ViDiC-1K with 1,000 video pairs by aggregating data from existing public sources while also generating videos via our proprietary pipeline. The approximate proportions of specific data sources are shown in Figure~\ref{fig:right}. To maintain a high standard of data quality, all videos were uniformly filtered to remove duplicates, videos containing significant artifacts, videos with negligible motion, and those exhibiting excessively large inter-video differences.

\begin{figure}[htbp]
    \centering
    \begin{minipage}[b]{0.48\textwidth}
        \centering
        \includegraphics[width=\textwidth, height=0.15\textheight, keepaspectratio]{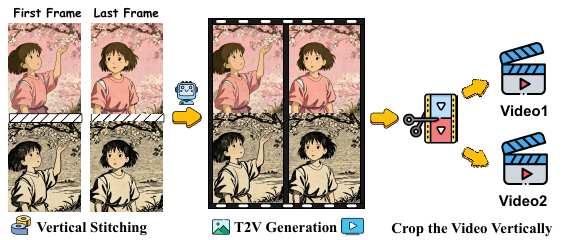}
        \vspace{0.5mm}
        \caption{Synthetic Generation via Frame Splicing.}
        \label{fig:videogenerate_1}
    \end{minipage}
    \hfill
    \begin{minipage}[b]{0.48\textwidth}
        \centering
        \includegraphics[width=\textwidth, height=0.15\textheight, keepaspectratio]{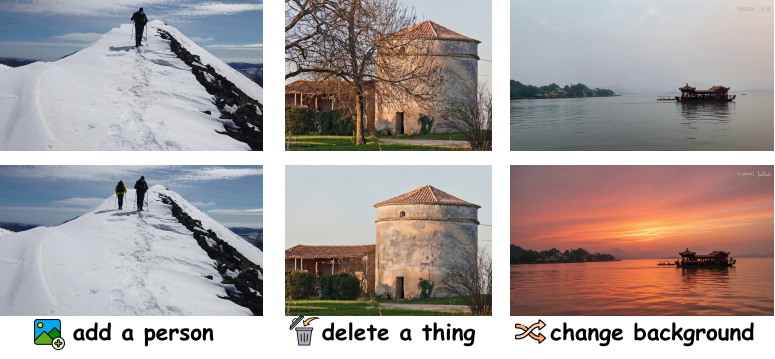}
        \caption{Edit via CV and Rendering tools.}
        \label{fig:videogenerate_2}
    \end{minipage}
\end{figure}

\begin{itemize}
\item \textbf{Externally Sourced Video Collection:}
The external data was sourced from two primary channels: public academic datasets (VidDiffBench \citep{burgess2025video}, IF-VidCap~\citep{li2025ifvidcapvideocaptionmodels}, VACE~\citep{vace}, PKU-DyMVHumans \citep{zheng2024PKU-DyMVHumans}, ToCaDa~\citep{malon2018toulouse} and DVSC~\citep{yokoo2024vsc2022}) and web platforms (YouTube, and LMArena \citep{zheng2023judging}). For the IF-Vidcap dataset and a YouTube subset, we employ a temporal bisection strategy, selecting continuous long takes and manually dividing each into two consecutive segments to generate similar video pairs.

\item \textbf{Controlled Synthetic Generation via Frame Splicing:}
To acquire video samples for categories that are difficult to collect naturally, such as dynamic weather transitions, we designed the synthetic pipeline illustrated in Figure \ref{fig:videogenerate_1}. The core process involves stacking boundary frames, utilizing the Veo3 model \citep{wiedemer2025videomodelszeroshotlearners} to synthesize a composite video, and subsequently splitting it. This synthetic data constitutes only a small fraction of our self-collected dataset. The specific prompts and detailed generation workflow are provided in the supplementary material.

\item \textbf{CV and Rendering-Based Video Augmentation:} As illustrated by the examples in Figure~\ref{fig:videogenerate_2}, we utilize a variety of computer vision and rendering techniques to achieve fine-grained editing of video content. This approach allows for precise modifications, including: (1) altering camera perspectives via ReCamMaster~\citep{bai2025recammaster}; (2) modifying artistic styles with stylization tools~\citep{ye2025stylemaster}; (3) adding or removing specific subjects using SAM-2~\citep{ravi2024sam2} combined with inpainting; and (4) re-animating subject actions within rendering engines such as Unreal Engine.

\end{itemize}

\subsubsection{Annotation Pipeline}
\label{sec:annotation}

Our annotation pipeline employs a two-stage process that combines automated generation with expert validation to ensure quality.

\noindent\textbf{Stage 1: Automated Draft Generation.}
This stage begins with the generation of comparative descriptions for video pairs. To mitigate single-model bias and diversify linguistic patterns, we distribute this description task across a pool of MLLMs: GPT-5~\citep{openai2025gpt5}, Gemini-2.5-Pro~\citep{gemini2.5}, Qwen3-VL-plus~\citep{qwen3}, InternVL-3.5-241B~\citep{internvl3.5}, and Doubao-Seed-1.8\citep{seed2025}. These models are explicitly prompted to prioritize high-level semantic discrepancies over low-level pixel artifacts. This constraint ensures that the generated descriptions capture meaningful content variations rather than technical imperfections. Subsequently, we employ a diverse set of models, including Gemini-2.5-Pro, GPT-5-mini~\citep{openai2025gpt5}, and DeepSeek-V3~\citep{deepseekv3}, to synthesize these insights into a unified and robust checklist.

\noindent\textbf{Stage 2: Human Validation.}
A team of six trained professional annotators meticulously refined the draft checklists. Each list was independently reviewed and corrected by two annotators based on unified criteria, targeting issues such as factual errors, logical contradictions, misclassifications, or excessive subjectivity. Crucially, the review scrutinized semantic validity, explicitly discarding items that were devoid of meaningful content or involved visual distinctions too subtle for humans to perceive. Any disagreements were resolved through a consensus-driven discussion mediated by a third senior annotator. This rigorous, multi-annotator protocol resulted in only 16.32\% of the initial model-generated items being retained verbatim; the remainder were either substantially revised or discarded entirely. This process ensures that every item in the final checklists is factually accurate, and precisely aligned with human judgment.

\subsection{Dataset Statistics}
\paragraph{\textbf{Overall Statistics}}

Our benchmark comprises 1,000 video pairs annotated with 3720 comparative checklist items (982 similarity, 2738 difference). Figure~\ref{fig:middle_top} illustrates the number of video pairs corresponding to each checklist length. The source videos are curated for diversity, with durations predominantly ranging from 2 to 12 seconds, reflecting the typical length observed in modern video editing(Figure~\ref{fig:duration}), a varied set of resolutions (Figure~\ref{fig:resolution}), and a broad spectrum of topics to ensure generalizability (Figure~\ref{fig:left}).

Our comparative checklist items are categorized according to a multifaceted taxonomy, depicted in Figure~\ref{fig:top}. This system encompasses seven key dimensions: 1) Subject, covering its type, count, and detailed attributes from appearance to pose; 2) Style, utilizing a constrained list of objective descriptors (e.g., Anime, Oil Painting) to ensure consistency; 3) Background, describing background objects, weather, location, atmosphere, and lighting; 4) Camera Work, analyzing cinematographic elements like movement and scale; 5) Subject Motion, detailing action dynamics; 6) Positional Relationship, focusing on spatial arrangements; and 7) Playback Technique, identifying basic effects like slow-motion. Statistically, Subject, Background and Style account for approximately 57\% to align with mainstream editing types, while the 35\% allocated to dynamic categories emphasizes temporal changes, distinguishing ViDiC from IDC.

\paragraph{\textbf{Comparison with Other Benchmarks}}

Current benchmarks for visual comparison suffer from critical fragmentation, focusing either on static images or on isolated tasks. A comprehensive overview in Table~\ref{tab:benchmark_compare_final} highlights this gap. A critical comparison with the concurrent VidDiffBench~\citep{burgess2025video} further highlights the distinct value of ViDiC-1K. While VidDiffBench pioneers action differencing, it is strictly confined to skill assessment within five narrow domains (e.g., fitness, surgery), significantly limiting its applicability to identifying specific motion deviations. In contrast, ViDiC-1K captures a comprehensive spectrum of visual variations—including subject consistency, cinematography, and background changes—essential for universal video understanding. This breadth enables critical real-world applications beyond simple coaching, such as validating consistency in generative video editing, identifying video plagiarism for copyright protection, and analyzing complex scenes in intelligent surveillance. To overcome the domain-specific limitations of previous works, ViDiC-1K establishes a holistic framework with 35 fine-grained categories. Notably, regarding data scale, while VidDiffBench relies solely on 549 existing clips, we double the volume to 1,000 by integrating both diverse collected footage and self-produced samples. Finally, by benchmarking 17 diverse MLLMs—versus only five in VidDiffBench—ViDiC-1K serves as a robust, general-purpose foundation for the community.

\begin{table}[t]
\small
\centering
\caption{Comparison of benchmarks for image and video difference captioning and related tasks. We compare our ViDiC-1K with existing image difference captioning datasets, including Spot-the-Diff~\citep{jhamtani2018learning}, CLEVR-Change~\citep{Park2019RobustCC}, and OmniDiff~\citep{Liu2025OmniDiff}, as well as the video action differencing benchmark VidDiffBench~\citep{burgess2025video}. ``Syn.'' and ``Real'' denote synthetic and real-world data sources, respectively. The size denotes the number of samples in the test set. Reference-based evaluation is a metric that measures model output quality by comparison with pre-defined reference answers (e.g., BLEU, CIDEr).}
% \label{tab:benchmark_compare_final}
\label{tab:benchmark_compare_final}
\resizebox{\columnwidth}{!}{%
\begin{tabular}{lccccc}
\toprule
\textbf{Benchmark} & 
\textbf{Source} & 
\textbf{Task} &
\textbf{Category Count} &
\textbf{Size} &
\textbf{Evaluation} \\
\midrule
Spot-the-Diff & Real & Image Difference Captioning & 1 & 1,400 & Reference-based \\ 
CLEVR-Change & Syn. & Image Difference Captioning & 5 & 7,970 & Reference-based \\ 
OmniDiff & Real and Syn. & Image Difference Captioning & 12 & 1,560 & Reference-based \\ 
VidDiffBench & Real & Video Action Differencing & 5 & 549 & Ground Truth + LLM \\ 
\midrule 
\textbf{ViDiC-1K (Ours)} & Real and Syn. & Video Difference Captioning & 35 & 1,000 & Checklist + LLM \\
\bottomrule
\end{tabular}}

\end{table}

\subsection{Evaluation Methodology}

\subsubsection{Evaluation Framework}
\label{sec:eval_framework}

Traditional metrics fall short of the evaluation for complex descriptive tasks, as they measure textual similarity rather than factual correctness. To overcome this, we propose a framework to directly quantify factual accuracy using a human-annotated checklist. This checklist is composed of a set of binary (yes/no) questions, denoted as $\mathcal{Q}$, derived from predefined evaluation dimensions. Each question has a corresponding ground-truth answer, forming a ground-truth answer set, $\mathcal{A}_{GT}$. During evaluation, the model under review $\mathcal{M}$, is prompted with a given video pair and the evaluation dimensions to generate a description, $D$. Subsequently, a powerful and separate Judge model, $\mathcal{J}$ (we use GPT-5-Mini), must answer the questions in $\mathcal{Q}$ based solely on the information present in $D$, without access to the videos. This process yields the Judge's answer set, $\mathcal{A}_{\mathcal{J}}$. The factual accuracy of the description is then determined by the consistency between $\mathcal{A}_{\mathcal{J}}$ and $\mathcal{A}_{GT}$, providing a direct and reliable measure of the model's ability to articulate facts.

The rationale for our evaluation method is twofold. First, we align with the binary judgment paradigm validated by MME\citep{fu2025mmecomprehensiveevaluationbenchmark}, POPE\citep{li2023evaluatingobjecthallucinationlarge}, and HallusionBench\citep{guan2024hallusionbenchadvanceddiagnosticsuite}, which has proven effective in probing fine-grained details and diagnosing hallucinations. Second, to strictly prevent information leakage, the model under test is exposed exclusively to the open-ended prompts and remains blind to the specific judgment questions or verification criteria.

\subsubsection{Evaluation Metric}
\label{sec:eval_metric}
 
We formulate the fine-grained video comparison metric as the Accuracy over the question set $\mathcal{Q}$:

\vspace{-1mm}
\begin{equation} \label{eq:accuracy}
\text{Accuracy} = \frac{1}{|\mathcal{Q}|} \sum_{i=1}^{|\mathcal{Q}|} \mathbb{I}\big(\mathcal{A}_{\mathcal{J},i} = \mathcal{A}_{GT,i}\big)
\end{equation}
\vspace{-1mm}

Here, $\mathbb{I}(\cdot)$ is an indicator function that equals 1 if the model's answer for question $i$ exactly matches the ground-truth answer, and 0 otherwise. Given the distinct design objectives of similarity and difference questions, we adopt tailored evaluation strategies for each type.

\noindent\textbf{Similarity Questions}
To penalize hallucination over omission (since enumerating all shared attributes in similar video pairs is impractical), questions about similarities are framed inversely (See Figure~\ref{fig:page} for an example.). A response is considered correct if it either confirms the similarity or omits the attribute, thereby only penalizing hallucinated differences.

\noindent\textbf{Difference Questions}
Conversely, to enforce descriptive accuracy, Difference questions are framed as verifiable propositions about specific differences (See Figure~\ref{fig:page} for an example). The model must correctly affirm these true statements, with any failure to verify or omission of the specified details being penalized.

\subsection{Training Dataset Construction}

To enhance the model's capability in video difference captioning, we constructed a training dataset comprising over \textbf{60k video pairs}. The construction methodology mirrors that of our test set; therefore, we omit repetitive descriptions here.

\textbf{Data Sourcing and Filtering.} Our data is aggregated from six diverse sources: Ditto-1M\citep{bai2025scalinginstructionbasedvideoediting}, LMArena, Miradata\citep{ju2024miradatalargescalevideodataset}, Ego-Exo4D\citep{grauman2024egoexo4dunderstandingskilledhuman}, MultiCamVideo\citep{bai2024syncammaster}, and Vript\citep{yang2024vriptvideoworththousands}. Before processing, we utilized hash mapping to identify and remove any samples overlapping with the evaluation set. Subsequently, we implemented a two-stage filtering pipeline to ensure data quality. First, we applied heuristic filters based on hard metrics, calculating blurriness via Laplacian variance and motion dynamics via optical flow magnitude to remove low-quality samples. Second, specifically for datasets exhibiting large internal quality variance, we utilized Qwen3-VL-32B to perform a secondary semantic screening.

\textbf{Similarity and Difference Annotation.} We employed Qwen3-VL-32B to generate training targets that explicitly capture semantic similarities and differences. This model was selected for its top-tier performance on our benchmark and its open-source availability, enabling high-quality annotations at lower cost.

Comprehensive details regarding the specific video construction strategies for each video source, the quality screening prompts, and the annotation prompts are provided in the supplementary material.

\section{Experiments}
\label{sec:experiments}

\subsection{Main Results}

We evaluate 17 popular models including Gemini-3.0-Flash, Gemini-2.5-Pro~\citep{gemini2.5}, Gemini-2.5-Flash~\citep{gemini2.5}, GPT-5~\citep{openai2025gpt5}, InternVL3.5~\citep{internvl3.5}, Qwen2.5-VL~\citep{qwen2.5vl}, Qwen3-VL~\citep{qwen3}, Keye-VL-1.5~\citep{kwaikeyeteam2025kwaikeyevltechnicalreport}, Mimo-VL-SFT~\citep{coreteam2025mimovltechnicalreport}, Kimi-VL-A3B~\citep{kimiteam2025kimivltechnicalreport}, GLM-4.1V~\citep{glmvteam2025glm41vthinkingversatilemultimodalreasoning}, InternVideo2.5~\citep{wang2025internvideo} and LLaVA-v1.6-Vicuna~\citep{liu2023improvedllava}. To establish a human baseline, we also invited independent evaluators unaffiliated with this project to participate in the evaluation, resulting in a human performance score. Additionally, we fine-tune Qwen2.5-VL-7B-Instruct on our self-constructed dataset to create \textbf{ViDiC-Qwen}. The main results are presented in Table~\ref{tab:mainresults_detailed}, which lead to the following key observations:

\begin{enumerate}
    \item \textbf{Model Performance Gap.} Our dataset reveals a clear performance hierarchy among models. While proprietary models still lead, open-source models like Qwen3-VL-32B are now outperforming some closed-source rivals, demonstrating rapid progress. Additionally, performance consistently scales with model size within a given model family.
    
    \item \textbf{Semantic Dimension Variations.} Models excel at Style recognition and perform reasonably on Subject, Position, and Background. However, Motion, Camera work and Playback technique detection remain particularly weak, especially for open-source models, indicating critical limitations in temporal artifact identification.
    
    \item \textbf{Similarity and Difference Trade-off.} High Similarity scores indicate low hallucination, but low Difference scores reveal weak fine-grained perception. Qwen3-VL-8B$^{\text{\tiny\faLightbulb}}$ achieves 80.24\% on Similarity but only 49.43\% on Difference, capturing coarse distinctions while missing subtle details. Balancing both remains a critical challenge.
    
    \item \textbf{Thinking Mode Impact.} Thinking mode improves both Difference and Similarity scores, revealing enhanced fine-grained perception and reduced hallucinations on identical content.

    \item \textbf{Incompatibility with Dual-Video Inputs.} Furthermore, MLLMs like LLaVA-v1.6-Vicuna-7B exhibited pathological behaviors on dual-video inputs, such as generating repetitive, non-terminating text.

    \item \textbf{Training Effectiveness.} 
    Our trained model achieves a remarkable average improvement of \textbf{11.75 points} (50.43 vs. 38.68). This significant boost effectively verifies the effectiveness of our constructed training set. We also further evaluated the model on external benchmarks. As shown in Table \ref{tab:right}, it outperforms the baseline on LVBench\citep{wang2025lvbenchextremelongvideo} (38.67 vs. 35.64), with improvements across specific fine-grained categories. Evaluation results on other benchmarks are provided in the supplementary material.

\end{enumerate}

\begin{table*}[!t]
\caption{
    Results of different models on our benchmark across overall metrics and fine-grained categories.
    \textbf{Diff.}, \textbf{Sim.}, and \textbf{Avg.} stand for Difference, Similarity, and Average scores(all in \%).
    \textbf{Pos.}, \textbf{Backgr.}, and \textbf{Tech.} denote Position, Background, and Playback Technique, respectively(all in \%).
    \textbf{Param.} denotes the parameter scale.
    A superscript lightbulb icon ({\text{\faLightbulb}}) indicates a ``thinking'' mode.
}
\centering
\small
\setlength{\tabcolsep}{4pt}
\begin{tabular}{l|c|ccc|ccccccc}
\toprule
\multirow{2}{*}{\textbf{Model}} & \multirow{2}{*}{\textbf{Param.}} & \multicolumn{3}{c|}{\textbf{Overall Metrics}} & \multicolumn{7}{c}{\textbf{Category Performance}} \\
\cmidrule(lr){3-5} \cmidrule(lr){6-12}
& & \textbf{Avg.}& \textbf{Diff.} & \textbf{Sim.}& \textbf{Subject} & \textbf{Motion} & \textbf{Pos.} & \textbf{Backgr.} & \textbf{Cam.} & \textbf{Style} & \textbf{Tech.} \\ 
\midrule
Human & \faBrain & 94.46 & 92.99 & 98.57 & 96.36 & 94.36 & 90.14 & 96.70 & 92.90 & 82.31 & 97.18\\
\midrule
\multicolumn{12}{c}{\textit{Closed-Source}}\\ 
\midrule
Gemini-2.5-Pro          & \faLock & \textbf{69.33}& \textbf{66.84}& 76.27& \textbf{71.95} & \textbf{61.71} & \textbf{70.42}& \textbf{75.47}& 60.41& 79.27& \textbf{66.20}\\
Gemini-3.0-Flash        & \faLock & 65.81& 60.04& 81.87& 66.17& 57.78& 68.31& 69.31& \textbf{63.88}& 77.44& 61.97\\
Gemini-2.5-Flash        & \faLock & 63.73& 57.87 & 80.04& 66.60& 56.92& 64.79& 66.12& 58.99& \textbf{81.71}& 42.25\\
GPT-5                   & \faLock & 62.26 & 62.03 & 62.90 & 62.63 & 56.79 & 68.05 & 74.03 & 49.75 & 61.18 & 40.62\\
\midrule
\multicolumn{12}{c}{\textit{Open-Source}} \\
\midrule
Qwen3-VL& 32B& 63.90 & 62.75 & 67.11 & 66.64 & 55.38 & 69.01 & 70.30 & 58.20 & 62.20 & 45.07\\ 
Qwen3-VL$^{\text{\tiny\faLightbulb}}$& 8B  & 57.57 & 49.43 & 80.24 & 59.70 & 48.03 & 59.86 & 62.27 & 54.73 & 71.95 & 26.76\\
Qwen3-VL& 8B  & 55.75 & 50.99 & 69.04 & 56.76 & 46.84 & 58.45 & 64.91 & 49.21 & 62.20 & 29.58\\
InternVL-3.5$^{\text{\tiny\faLightbulb}}$       & 38B  & 53.62 & 47.64 & 70.26 & 54.48 & 43.42 & 53.17 & 64.36 & 49.21 & 54.27 & 26.76\\
Mimo-VL-SFT$^{\text{\tiny\faLightbulb}}$& 7B  & 51.26 & 41.20 & 79.33 & 51.72 & 39.32 & 49.30 & 55.78 & 53.31 & 71.95 & 26.76\\
Qwen2.5-VL-Instruct & 72B  & 46.22 & 38.04 & 69.01 & 45.00 & 35.10 & 47.54 & 53.74 & 45.34 & 62.80 & 23.94\\
InternVL-3.5       & 38B  & 45.85 & 36.83 & 70.98 & 45.11 & 40.00 & 45.94 & 51.71 & 42.74 & 61.59 & 21.13\\
InternVL-3.5$^{\text{\tiny\faLightbulb}}$& 8B& 45.78 & 37.80 & 68.02 & 46.23 & 33.68 & 46.48 & 53.14 & 42.74 & 66.46 & 21.13\\
Qwen2.5-VL-Instruct& 32B& 45.30 & 35.55 & 72.48 & 45.28 & 35.62 & 46.83 & 52.53 & 43.92 & 52.44 & 22.54\\
Keye-VL-1.5$^{\text{\tiny\faLightbulb}}$& 8B & 45.24 & 30.94 & 85.12 & 43.99 & 35.89 & 45.55 & 50.72 & 45.76 & 63.98 & 21.43\\
Mimo-VL-SFT& 7B  & 43.09 & 33.27 & 70.47 & 45.67 & 32.82 & 43.31 & 44.88 & 45.58 & 51.83 & 22.54\\
Qwen2.5-VL-Instruct& 7B   & 38.68 & 25.90 & 74.31 & 35.95 & 32.53 & 35.21 & 41.52 & 43.44 & 57.32 & 22.54\\
InternVL-3.5& 8B   & 38.18 & 29.34 & 62.83 & 39.33 & 30.48 & 38.03 & 43.89 & 33.12 & 54.88 & 18.31\\
Keye-VL-1.5& 8B & 38.12 & 28.94 & 63.74 & 38.51 & 31.53 & 34.52 & 43.72 & 35.52 & 51.55 & 21.43\\
GLM-4.1V$^{\text{\tiny\faLightbulb}}$  & 9B  & 36.51 & 29.04 & 57.33 & 38.30 & 30.94 & 33.10 & 40.81 & 34.38 & 41.46 & 21.13\\
Kimi-VL-A3B$^{\text{\tiny\faLightbulb}}$ & 16B  & 34.82 & 21.23 & 72.71 & 33.21 & 28.03 & 31.34 & 35.97 & 40.69 & 52.44 & 21.13\\ 
InternVideo2.5   & 7B   & 34.18 & 16.95 & 82.26 & 29.76 & 30.60 & 32.75 & 33.00 & 42.74 & 57.32 & 21.13\\

% Llama-3.2& 11B & 20.46 & 0.95 & 74.85 & 15.38 & 21.03 & 20.77 & 11.99 & 38.33 & 28.66 & 21.13\\
LLaVA-V1.6-Vicuna & 7B & 25.19 & 0.58 & \textbf{93.79} & 17.89 & 25.98 & 22.18 & 17.16 & 43.69 & 49.39 & 22.54\\
\midrule
ViDiC-Qwen (Ours) & 7B & 50.43 & 41.72 & 74.69 & 50.37 & 38.70 & 52.11 & 57.38 & 48.73 & 68.10 & 26.76\\
\bottomrule
\end{tabular}
\label{tab:mainresults_detailed}
\end{table*}

\subsection{Further Analysis}

\paragraph{\textbf{Judge Consistency Analysis.}}

To select the most suitable LLM as an automated judge, we conducted a human–model inter-rater reliability analysis. To balance efficiency and reliability, we randomly sampled 750 video pairs, accounting for 75\% of our dataset. For this subset, we aggregated responses generated by multiple models, including Gemini-2.5-pro~\citep{gemini2.5}, Qwen2.5-VL~\citep{qwen2.5vl}, and GPT-5~\citep{openai2025gpt5}. These responses were then independently assessed by both human annotators (serving as the baseline) and three candidate LLM judges: GPT-5 Mini~\citep{openai2025gpt5}, DeepSeek-V3~\citep{deepseekv3}, and Qwen3-32B~\citep{qwen3}.  The concordance rates, which quantify the alignment of each LLM's judgments with the human evaluation standards, are summarized in Table \ref{tab:left}. The results indicate a strong correlation, particularly for GPT-5 Mini, validating the potential of using LLMs for scalable and consistent evaluation. This is also the reason why we select GPT-5 Mini.

To ensure reliability and rule out stochastic fluctuations, we conducted five repeated judging rounds using GPT-5-mini on captions generated by Gemini-2.5-Pro. The accuracy remained stable (69.30\%–69.62\%) with high human agreement (94.38\%–95.42\%), confirming the framework's robustness.

\begin{table}[t]
    \centering
    \renewcommand{\arraystretch}{1.35}
    \caption{Concordance Rates of LLM Judges with a Human Baseline. The table compares agreement percentages.}
    \label{tab:concordance}
    \begin{tabular}{lccc}
        \toprule
        \textbf{LLMs} & \textbf{Average} & \textbf{Similarity} & \textbf{Difference} \\
        \midrule
        GPT-5-mini  & 95.22 & 95.90 & 94.97 \\
        DeepSeek-V3 & 89.37 & 90.84 & 88.84 \\
        Qwen3-32B   & 87.23 & 88.98 & 86.60 \\
        \bottomrule
    \end{tabular}
\end{table}

\begin{table}[t]
    \centering
    \renewcommand{\arraystretch}{1.35}
    \caption{Comparison on fine-grained metrics. Baseline denotes Qwen2.5VL-7B; Ours denotes ViDiC-Qwen-7B.}
    \label{tab:finegrained}
    \begin{tabular}{l|cccc}
        \toprule
        Model & 
        Entity Recognition & 
        Temporal Grounding & 
        Summarization & 
        Key Info Retrieval \\
        \midrule
        Baseline & 32.35 & 31.36 & 31.03 & 36.77 \\
        \textbf{Ours} & \textbf{37.96} & \textbf{36.82} & \textbf{34.48} & \textbf{39.18} \\
        \bottomrule
    \end{tabular}
\end{table}

% \begin{table}[t]
% \centering
% \caption{Detailed performance comparison on LVBench.}
% \label{tab:lvbench}
% % 如果表格太宽超出页面，可以取消下面这行 resizebox 的注释
% % \resizebox{\linewidth}{!}{
% \begin{tabular}{l|c|cccccc}
% \toprule
% \textbf{Model} & \textbf{Overall} & \textbf{Key Info} & \textbf{Event} & \textbf{Summ.} & \textbf{Entity} & \textbf{Reasoning} & \textbf{Temporal} \\
% \midrule
% Baseline & 35.64 & 36.77 & 36.48 & 31.03 & 32.35 & 38.81 & 31.36 \\
% \textbf{Ours} & \textbf{38.67} & \textbf{39.18} & \textbf{38.49} & \textbf{34.48} & \textbf{37.96} & \textbf{39.80} & \textbf{36.82} \\
% \bottomrule
% \end{tabular}
% % }
% \end{table}

\paragraph{\textbf{Effect of Different Video Parameters.}}

\begin{figure*}[!t]
    \centering
    \setlength{\belowcaptionskip}{-3pt} % 调整主标题与下方内容的间距，如果需要的话
    \setlength{\abovecaptionskip}{6pt}  % 调整子图与主标题的间距，如果需要的话
    % Left Figure Block (contains two stacked subfigures)
    \subcaptionbox{Comparison of accuracy for two models: Mimo-VL-SFT$^{\text{\tiny\faLightbulb}}$ and Qwen2.5-VL-7B across varying fps and resolutions.\label{fig:frame_resolution_accuracy}}{%
        \begin{minipage}[b]{0.3\textwidth}
            \includegraphics[width=\linewidth]{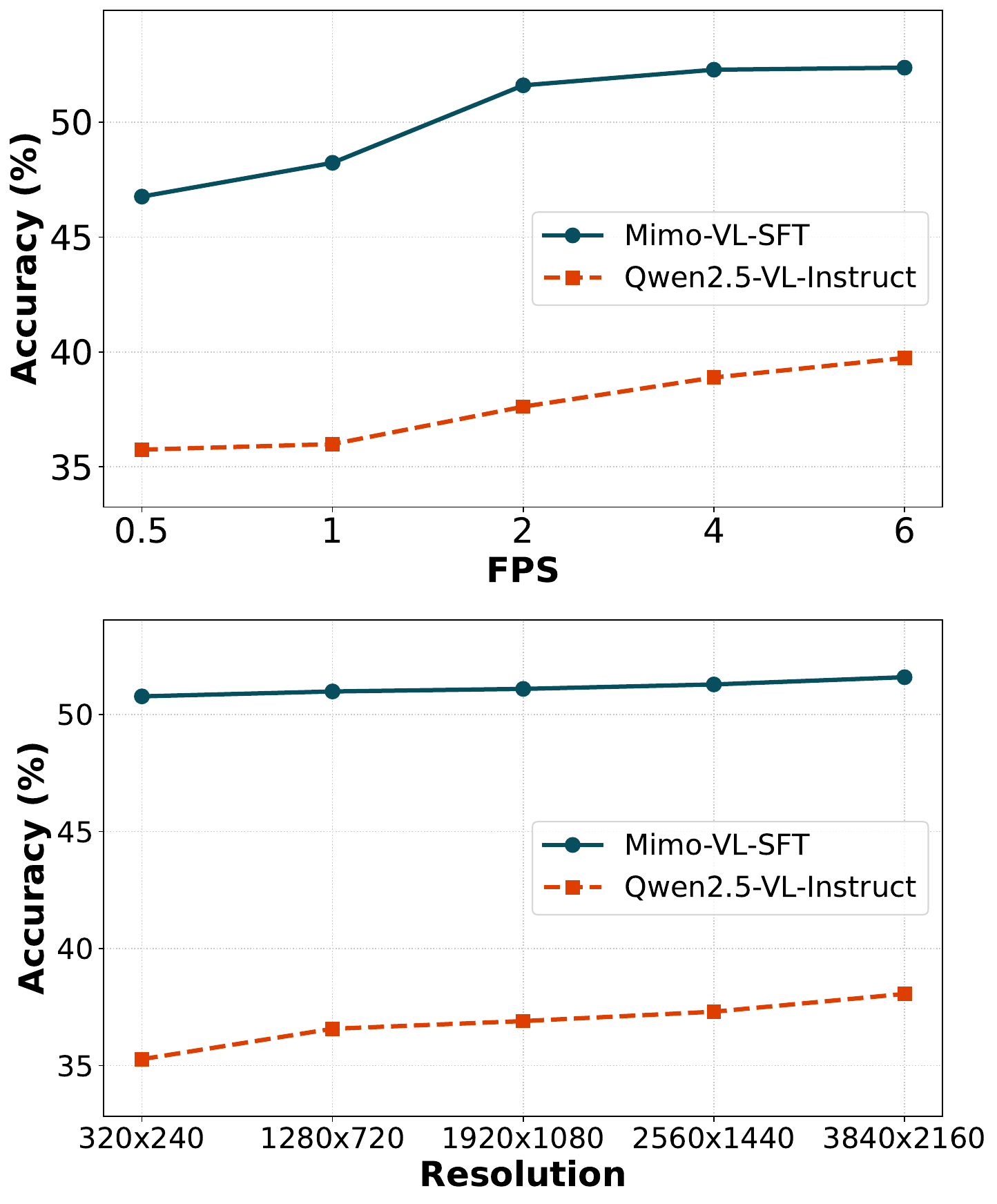}
        \end{minipage}%
    }
    \hfill % Horizontal space between figures
    % Middle Figure Block
    \subcaptionbox{Three video augmentations (blur, noise, color saturation) applied at light, medium, and heavy intensities, showing their progressive effects on frames.\label{fig:video_change}}{%
        \begin{minipage}[b]{0.32\textwidth}
            \centering
            \includegraphics[width=\linewidth]{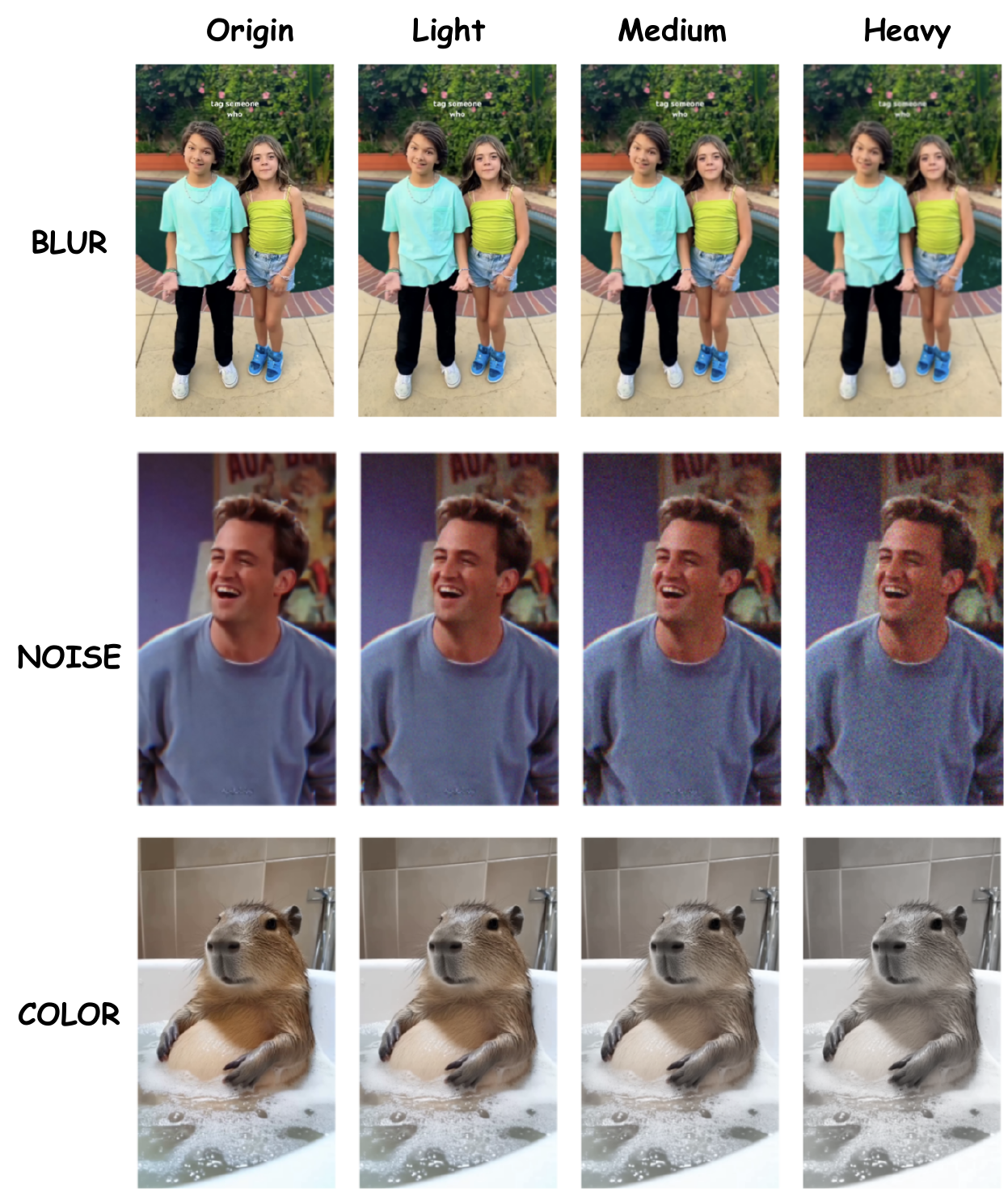}
        \end{minipage}%
    }
    \hfill % Horizontal space between figures
    % Right Figure Block
    \subcaptionbox{Difference detection scores across augmentation types and intensity levels for Qwen2.5-VL-7B (top) and Qwen2.5-VL-32B (bottom).\label{fig:augmentation_comparison}}{%
        \begin{minipage}[t]{0.32\textwidth} % 改为顶部对齐
            \centering
            \includegraphics[width=\linewidth]{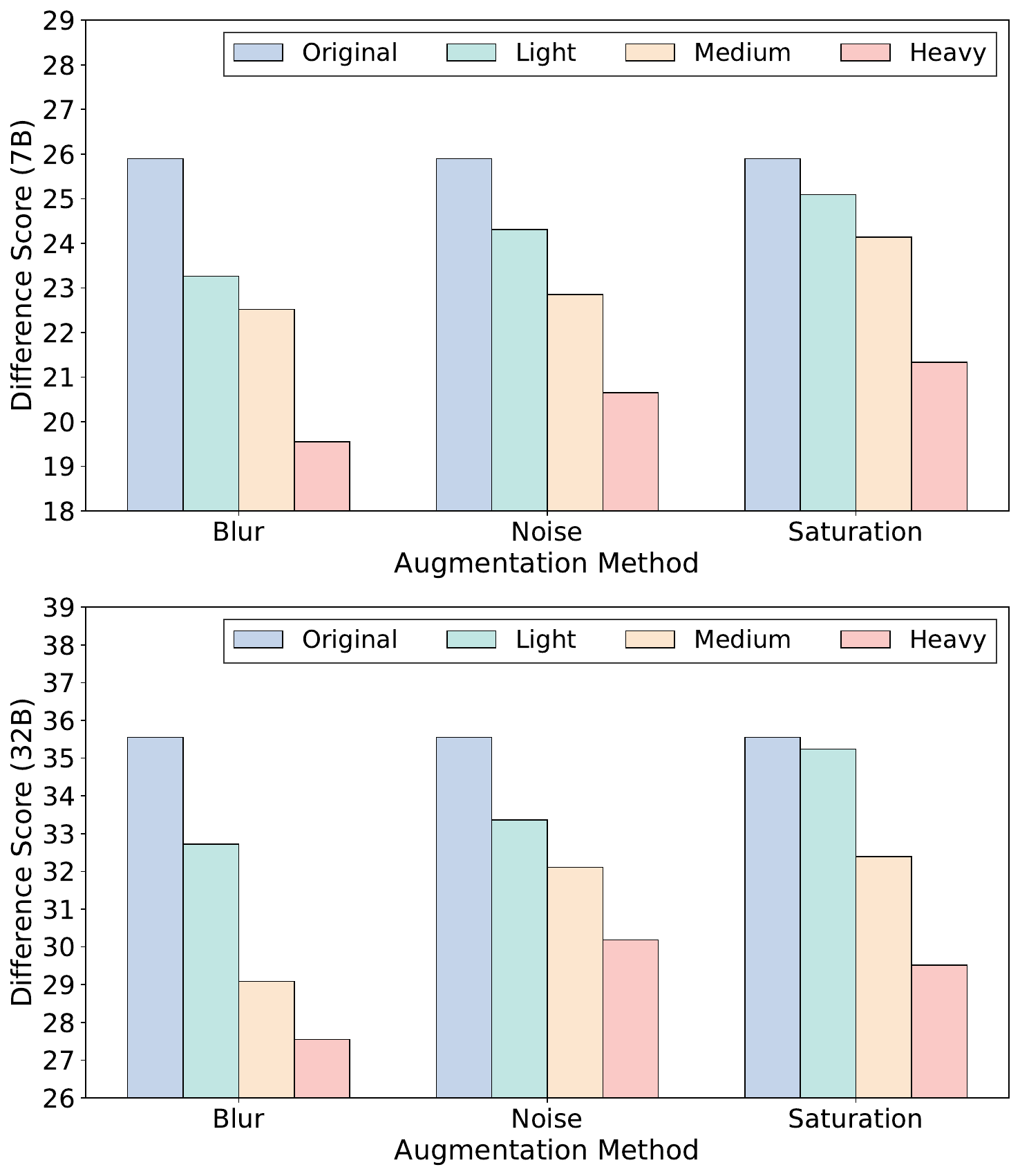}
        \end{minipage}%
    }
    \caption{Overview of video parameters and data augmentation effects. (a) Model accuracy with respect to fps and resolutions. (b) Examples of video augmentations at different intensity levels. (c) Quantitative impact of augmentation intensity on model performance metrics.}
    \label{fig:combined_overview}
\end{figure*}

We conducted a sensitivity analysis on Qwen2.5-VL-7B-Instruct and Mimo-VL-SFT$^{\text{\tiny\faLightbulb}}$ to evaluate the impact of temporal sampling and spatial resolution. To isolate these effects, we fixed the frame rate at 2 fps while varying the resolution, and fixed the resolution at $720 \times 1280$ while varying the FPS. As illustrated in Figure~\ref{fig:frame_resolution_accuracy}, accuracy for both models exhibits a continuous upward trend across both dimensions. This outcome is expected, as increasing data density in both time and space provides more comprehensive visual cues for accurate captioning. However, the magnitude of improvement differs significantly. The performance gain driven by FPS is much steeper than that of resolution. Notably, Mimo-VL-SFT$^{\text{\tiny\faLightbulb}}$ suffers a significant performance drop at the extremely low sampling rate of 0.5 fps, indicating a critical threshold for temporal information. In comparison, while higher resolutions yield consistent gains, the growth rate is notably smaller and more gradual relative to FPS. This suggests that Video Difference Captioning tasks depend more heavily on motion continuity than fine-grained spatial details, distinguishing Video Difference Captioning from Image Difference Captioning (IDC), which typically rely more on spatial fidelity.

\paragraph{\textbf{Effect of Input Video Noise}}

To evaluate the robustness of VLMs against common visual corruptions, we conduct experiments on Qwen2.5-VL-7B and Qwen2.5-VL-32B models. We systematically apply blur, noise, and color saturation augmentations at three distinct intensity levels (light, medium, and heavy), visualized in Figure~\ref{fig:video_change}. Detailed augmentation parameters are provided in supplementary material. As shown in Figure~\ref{fig:augmentation_comparison}, the Difference scores exhibit a consistent decline as augmentation intensity increases across all distortion types. Notably, under light and medium augmentation conditions, the overall semantic content remains largely preserved, and the semantic differences between video pairs remain clearly perceivable to human observers. However, the models' Difference scores decrease substantially. We posit that this performance shift stems from a reduction in the model's fine-grained perceptual acuity; the introduced visual interference masks subtle, pixel-level details, thereby impairing its ability to discern minute distinctions. Through qualitative analysis of model outputs, we observe that augmented inputs lead to the omission of fine-grained details in the generated descriptions, such as gender attributes, clothing characteristics, and other nuanced visual elements. This suggests that while the models retain their capacity for high-level semantic understanding, their sensitivity to low-level visual features is significantly compromised under distortion.

\paragraph{\textbf{Order Sensitivity.}}
To investigate whether the order of video inputs affects model performance, we conduct experiments comparing two settings: Forward (inputting Video A followed by Video B) and Reverse (inputting Video B followed by Video A). We evaluate four representative models: Qwen3-VL-32B, Qwen2.5-VL-72B, InternVL3.5-38B$^{\text{\tiny\faLightbulb}}$, and Keye-VL-1.5-8B$^{\text{\tiny\faLightbulb}}$. Our results reveal distinct patterns of order sensitivity across different model architectures. The Qwen series demonstrates remarkable robustness to input sequence variations, with Qwen3-VL-32B achieving nearly identical performance (63.90 in Forward vs. 63.04 in Reverse) and Qwen2.5-VL-72B showing similar stability (46.22 vs. 45.45). In sharp contrast, InternVL3.5-38B$^{\text{\tiny\faLightbulb}}$ and Keye-VL-1.5-8B$^{\text{\tiny\faLightbulb}}$ exhibit notable sensitivity to input order. Specifically, InternVL3.5-38B$^{\text{\tiny\faLightbulb}}$ experiences a performance drop from 45.85 in the Forward setting to 42.66 in the Reverse setting, while Keye-VL-1.5-8B$^{\text{\tiny\faLightbulb}}$ declines from 45.24 to 41.67. We attribute this performance gap to differences in model architectures and attention mechanisms. The attention mechanisms in these susceptible models may exhibit an uneven focus on the visual inputs, disproportionately emphasizing one video while potentially \textbf{neglecting the other}, leading to inconsistent reasoning results when the video order is swapped.

\begin{figure*}[!t]
    \centering
    \includegraphics[width=\textwidth, keepaspectratio]{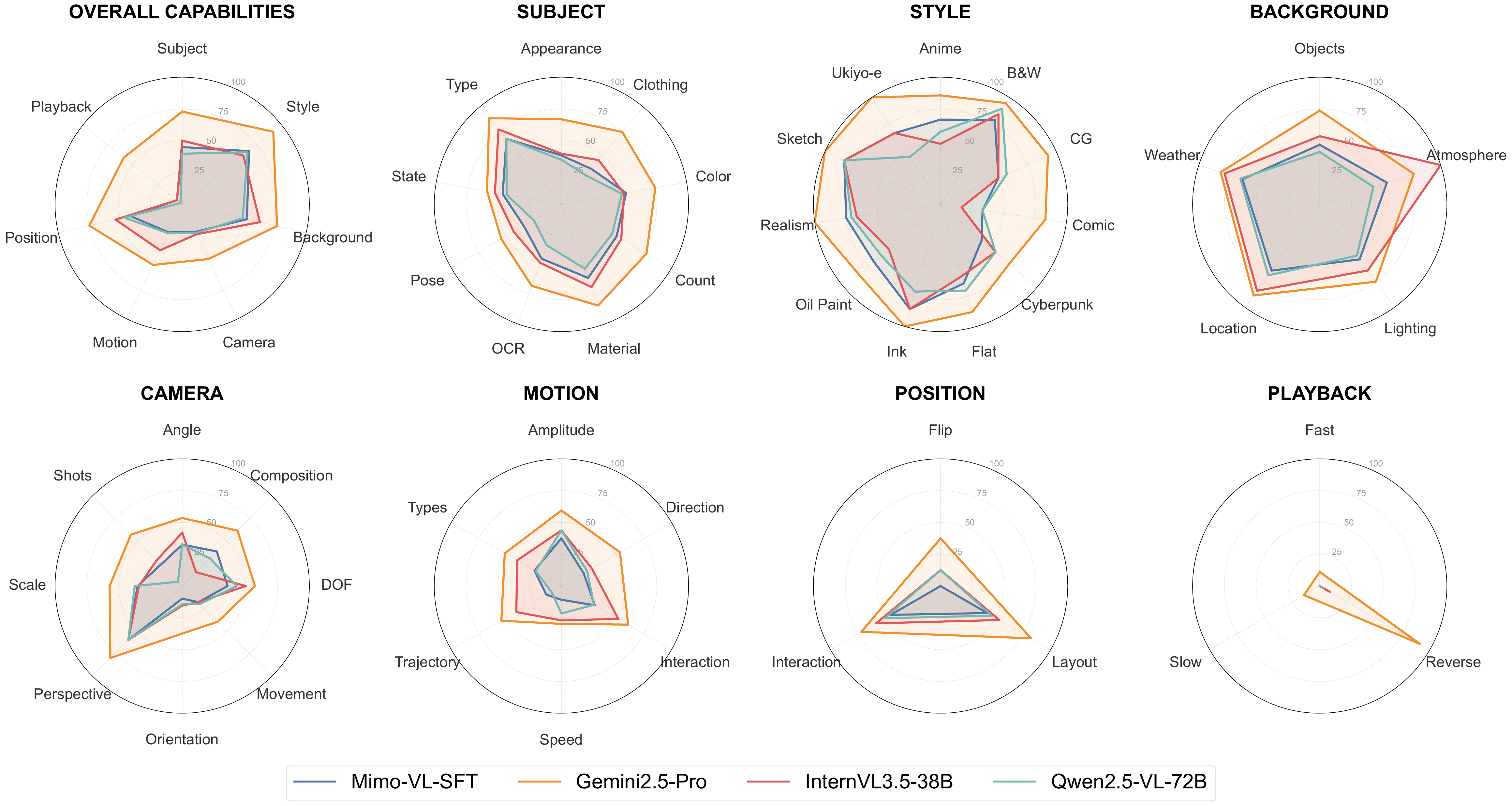}
    \caption{Detailed performance analysis. The top-left `Overall' chart summarizes model accuracy across seven high-level categories. The other seven charts offer a fine-grained breakdown for each of these categories, detailing performance on specific sub-tasks.}
    \label{fig:combined_radar_charts}
\end{figure*}

\paragraph{\textbf{Fine-Grained Category Accuracy Analysis.}}
To uncover distinct model capabilities often masked by aggregate-level analysis, we conducted a fine-grained evaluation of four representative multimodal models: Mimo-VL-SFT$^{\text{\tiny\faLightbulb}}$, Gemini-2.5-pro, InternVL3.5-38B$^{\text{\tiny\faLightbulb}}$, and Qwen2.5-VL-72B-Instruct. This diverse set, spanning proprietary and open-source multimodal models of varying scales, allows for a multifaceted comparison as illustrated in Figure~\ref{fig:combined_radar_charts}.

Although the models generally exhibit strong capabilities in recognizing core attributes such as Type, and Location, their performance on several fine-grained categories remains exceptionally challenging. These difficult areas reveal critical gaps in current model capabilities. Poor performance on tasks like OCR, combined 
with struggles in visual aspects such as expression and depth of field, not only differentiate models but also map out the precise frontiers where progress is most needed. By identifying these specific weaknesses, our 
analysis provides targeted guidance for future research priorities.

\begin{figure}[htbp]
\centering
\includegraphics[width=\textwidth, height=0.21\textheight, keepaspectratio]{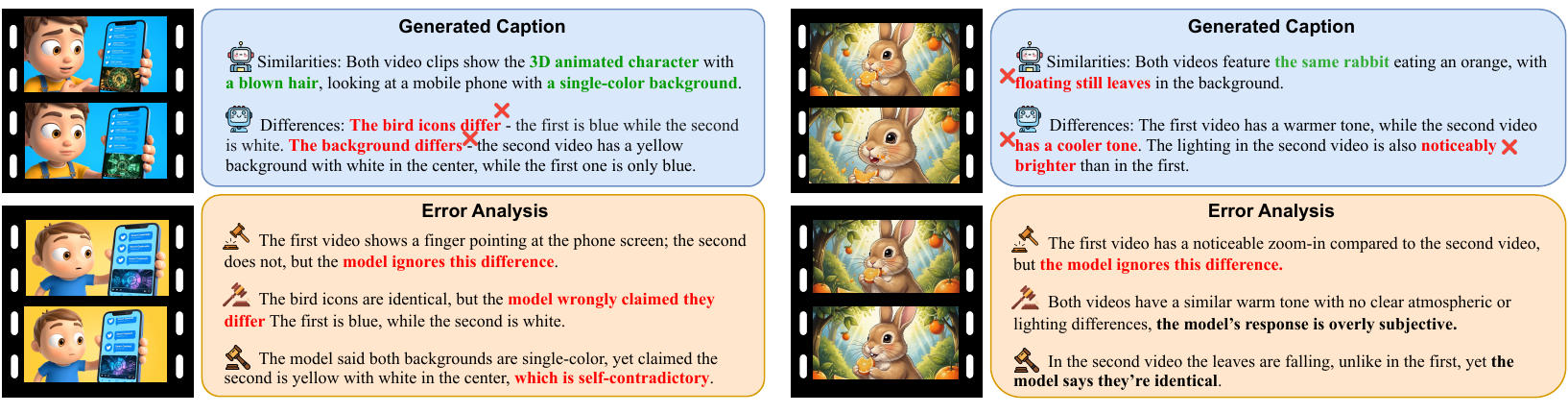}
\caption{We present two illustrative failure cases of the model. In each example, the left column displays two representative frames from the video pair. The top panel shows the generated caption, while the bottom panel identifies the errors and their underlying causes. For clarity, the model outputs have been simplified.}
\label{fig:error_analysis_combined}
\end{figure}

\paragraph{\textbf{Error Analysis.}}
An analysis of our model's failure cases reveals three recurring and distinct error patterns, as illustrated in Figure~\ref{fig:error_analysis_combined}: (1) hallucinating non-existent differences on identical content; (2) generating self-contradictory responses, simultaneously claiming sameness while describing a change; and most frequently, (3) incomplete or imprecise difference detection, where the model either fails to perceive a salient change or describes it in a vague, uninformative manner. Taking Gemini-2.5-Pro as a representative case, quantitative analysis shows that Missed Detection is the most prevalent error (577 cases, 50.6\%), followed by Self-Contradiction (323 cases, 28.3\%) and Hallucination (191 cases, 16.7\%). Detailed results for other models and experimental configurations are provided in the supplementary material. These errors collectively highlight current limitations in the model's fine-grained perception, logical consistency, and robust visual grounding, pointing to clear avenues for future improvement.

Furthermore, the rabbit example on the right of Figure~\ref{fig:error_analysis_combined} exposes a deeper, modality-specific challenge that distinguishes ViDiC from static Image Difference Captioning. In this scenario, the model fails to capture the temporal evolution of the scene, resulting in two critical omissions: it completely neglects the camera zoom-in operation in the first video and misinterprets the dynamic falling leaves in the second video as being stationary. These failures suggest that the model currently struggles to perceive changes that occur over time rather than just in visual appearance. This underscores the essential significance of the ViDiC task, as it demands a higher-level perception of temporal continuity and camera intent, capabilities that are indispensable for comprehensive video understanding but are absent in static image comparisons.

\section{Conclusion}
This paper introduces Video Difference Captioning (ViDiC), a novel task aimed at advancing comparative spatio-temporal video understanding. To support this task, we present a large-scale dataset comprising over 60,000 training samples and the ViDiC-1K benchmark, which spans seven critical dimensions. Furthermore, we propose the Dual-Checklist framework, an LLM-as-a-judge approach designed for rigorous and factually accurate assessment. Our evaluation of 17 representative multimodal models reveals significant bottlenecks in comparative perception, particularly in analyzing cinematography and motion. By highlighting these limitations, ViDiC provides a robust foundation and a new trajectory for developing fine-grained video understanding models.
\label{sec:conclusiton}

\section{Future Directions}

This work establishes a foundational benchmark for the novel task of Video Difference Captioning (ViDiC), introducing a high-quality test set (ViDiC-1K) and a dual-checklist evaluation framework that reveals the comparative spatio-temporal understanding capabilities of current MLLMs. Looking forward, we envision leveraging this methodology to advance video editing research. Specifically, our ViDiC framework can be adapted to construct large-scale, high-quality training datasets for video editing tasks. By systematically capturing fine-grained differences between original and edited videos, this approach can generate precise edit descriptions and annotations that significantly improve the quality of video editing training data. This direction holds promise for developing more capable video editing models that understand and execute complex editing instructions with greater accuracy and nuance.

\bibliographystyle{unsrtnat}
\bibliography{references} 

\clearpage
\setcounter{page}{1}
\appendix

\section{Real-world Applications of ViDiC}

ViDiC moves beyond simple similarity scoring to offer actionable, semantic insights into video discrepancies. By articulating what has changed, this capability empowers diverse domains:

\begin{itemize}[leftmargin=*, itemsep=0.2em]
    \item \textbf{Content Integrity and Forensics}: 
    ViDiC facilitates copyright protection by detecting subtle obfuscations. Furthermore, it combats disinformation by explicitly identifying and describing manipulated regions, acting as an interpretable tool for fact-checking.

    \item \textbf{Video Editing Verification}: 
    Serving as a rigorous evaluator, ViDiC validates whether editing models adhere to user prompts. It ensures precise modifications without introducing unintended artifacts or background alterations by explicitly captioning discrepancies between source and edited footage.

    \item \textbf{Automated Video Production}: 
    In collaborative workflows, the system streamlines post-production by generating automatic Change Logs. These summaries of editorial adjustments effectively bridge the communication gap between human editors and AI tools.

    \item \textbf{Skill Assessment and Rehabilitation}: 
    For applications in sports training or physical therapy, ViDiC provides granular feedback on pose deviations and temporal misalignments by comparing user performance against standard reference videos.

    \item \textbf{Scientific Monitoring}: 
    It automates change detection in longitudinal observation footage, converting visual shifts in dynamic environments into structured descriptive reports for efficient analysis.

    \item \textbf{Intelligent Surveillance}: Surpassing conventional motion sensing, semantic analysis distinguishes between environmental interference and relevant activity, ensuring that alerts are triggered only by specific objects or actionable security behaviors.
\end{itemize}

\section{Construction of the Training Set}

In this section, we detail the construction strategies employed for the diverse data sources within our training set. First, for the large-scale open-source datasets, we apply a quality filtering pipeline based on four key quantitative metrics: resolution, blurriness, brightness, and motion intensity. This rigorous screening removes samples with insufficient visual clarity, extreme lighting conditions, or lack of motion. Beyond these general metrics, we implement tailored strategies to accommodate the distinct properties of data sources. For Ditto-1M, which focuses on video editing, we aim to eliminate artifacts introduced during the editing process. We leverage the large multimodal model Qwen3-VL-32B to assess post-edit quality, filtering out videos that exhibit temporal instability or incoherence. Finally, for Vript and Miradata, we employ a long-shot segmentation strategy. By cutting continuous long-take footage into shorter segments, we successfully extract naturally cohesive and visually similar video pairs.

\begin{figure}[h]
    \centering
    \includegraphics[width=0.75\linewidth]{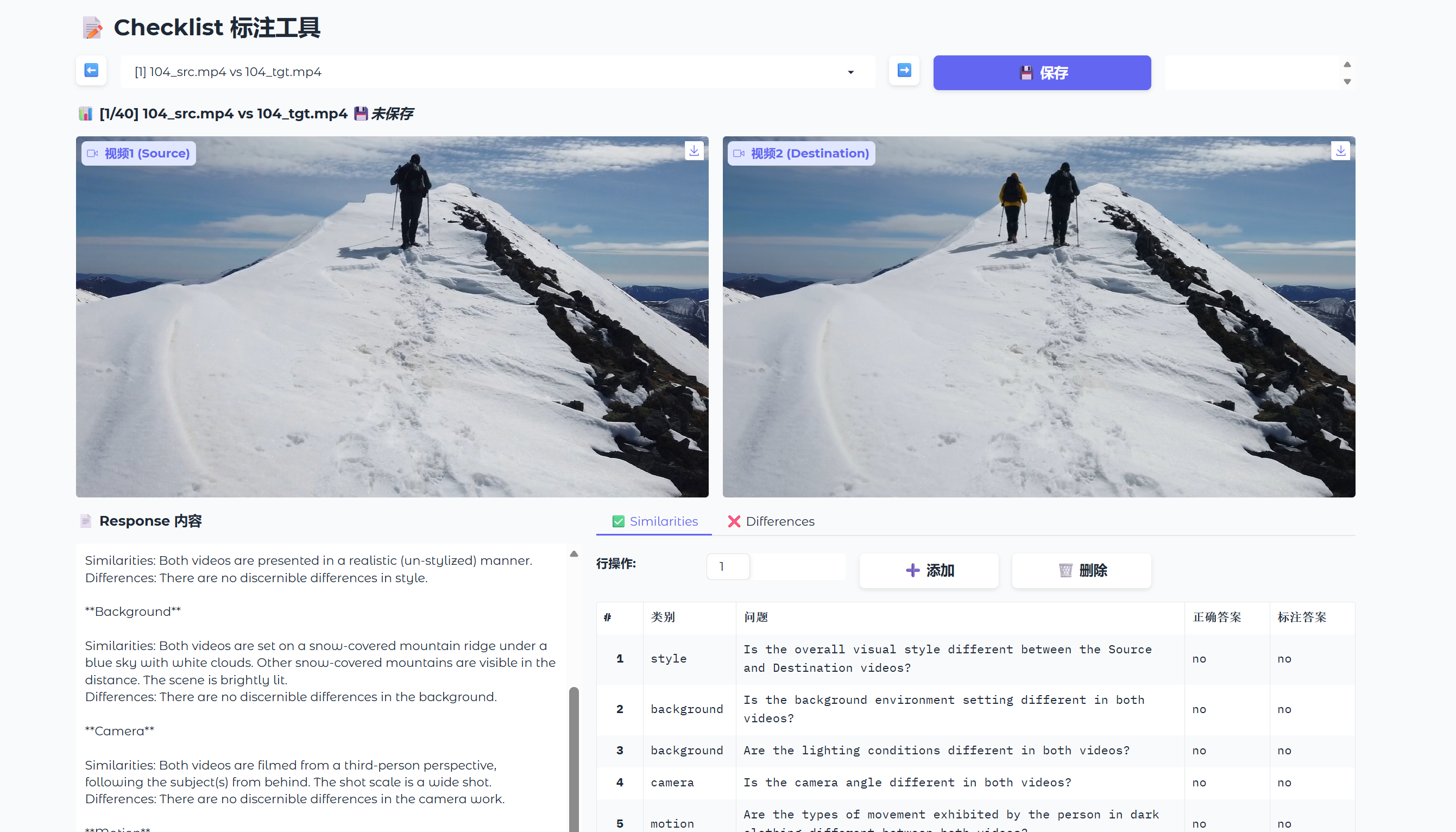} 
    \caption{The annotation interface used for the final inspection. Annotators utilize this tool to simultaneously filter out unqualified videos and validate the rationality of the checklist and standard answers.}
    \label{fig:annotation_interface}
\end{figure}

\section{Dataset Samples}
To better illustrate our dataset, this section presents two specific examples. Each example features key frames from a pair of similar videos, along with a corresponding checklist of questions.

\vspace{1mm}

\noindent
% Start a minipage for the LEFT column (Example 1)
\begin{minipage}[t]{0.48\linewidth}
    \footnotesize  % 改为更小的字体
    \textbf{Example 1}
    \begin{center}
        \includegraphics[width=1\linewidth]{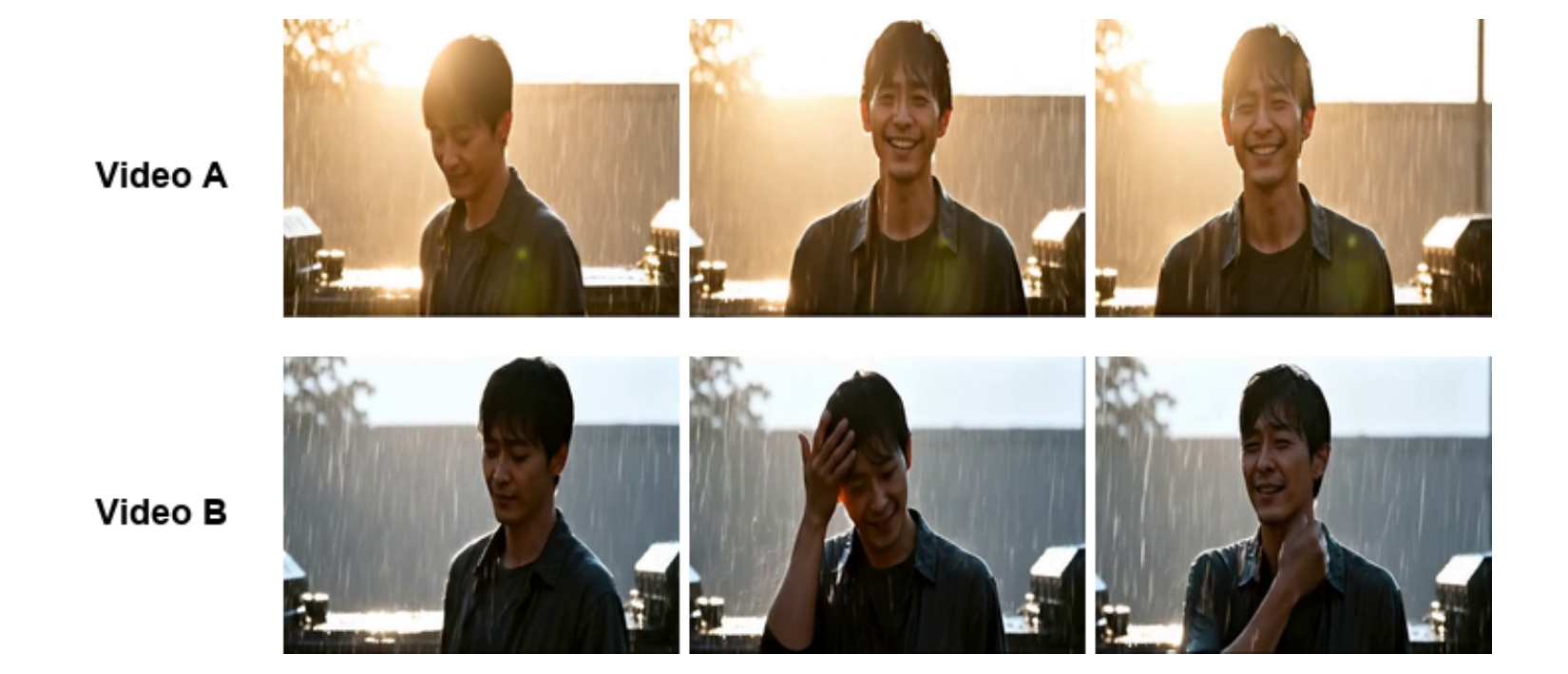}
    \end{center}

    \begin{tcolorbox}[mainbox, left=1mm, right=1mm, top=1mm, bottom=1mm]  % 减小内边距

        \begin{tcolorbox}[titlebox, colback=black!60]
        Similarity Question
        \end{tcolorbox}

        \begin{itemize}
        \item \textbf{Class:} Camera \\
        \textbf{Category:} Shot Scale \\
        \textbf{Question:} Is the camera's shot scale different in both videos? \\
        \textbf{Correct Answer:} no
            \vspace{0.2em}
        
        \item \textbf{Class:} Playback Technique \\
        \textbf{Question:} Is the playback technique different in both videos? \\
        \textbf{Correct Answer:} no
            \vspace{0.2em}
        \end{itemize}

        \begin{tcolorbox}[titlebox, colback=black!60]
        Difference Question
        \end{tcolorbox}

        \begin{itemize}
            \item \textbf{Class:} Motion \\
        \textbf{Category:} Motion Type \\
        \textbf{Question:} Does the man in the Video B touch his head with his hand, a motion that is not seen in the Video A? \\
        \textbf{Correct Answer:} yes
            \vspace{0.2em}

            \item \textbf{Class:} Background \\
        \textbf{Category:} Lighting \\
        \textbf{Question:} Is the lighting in the Video B white-toned, in contrast to the warm, golden-toned lighting in the Video A? \\
        \textbf{Correct Answer:} yes
            \vspace{0.2em}
        \end{itemize}

    \end{tcolorbox}
\end{minipage}%
\hfill
\begin{minipage}[t]{0.48\linewidth}
    \footnotesize  % 改为更小的字体
    \textbf{Example 2}
    \begin{center}
        \includegraphics[width=1\linewidth]{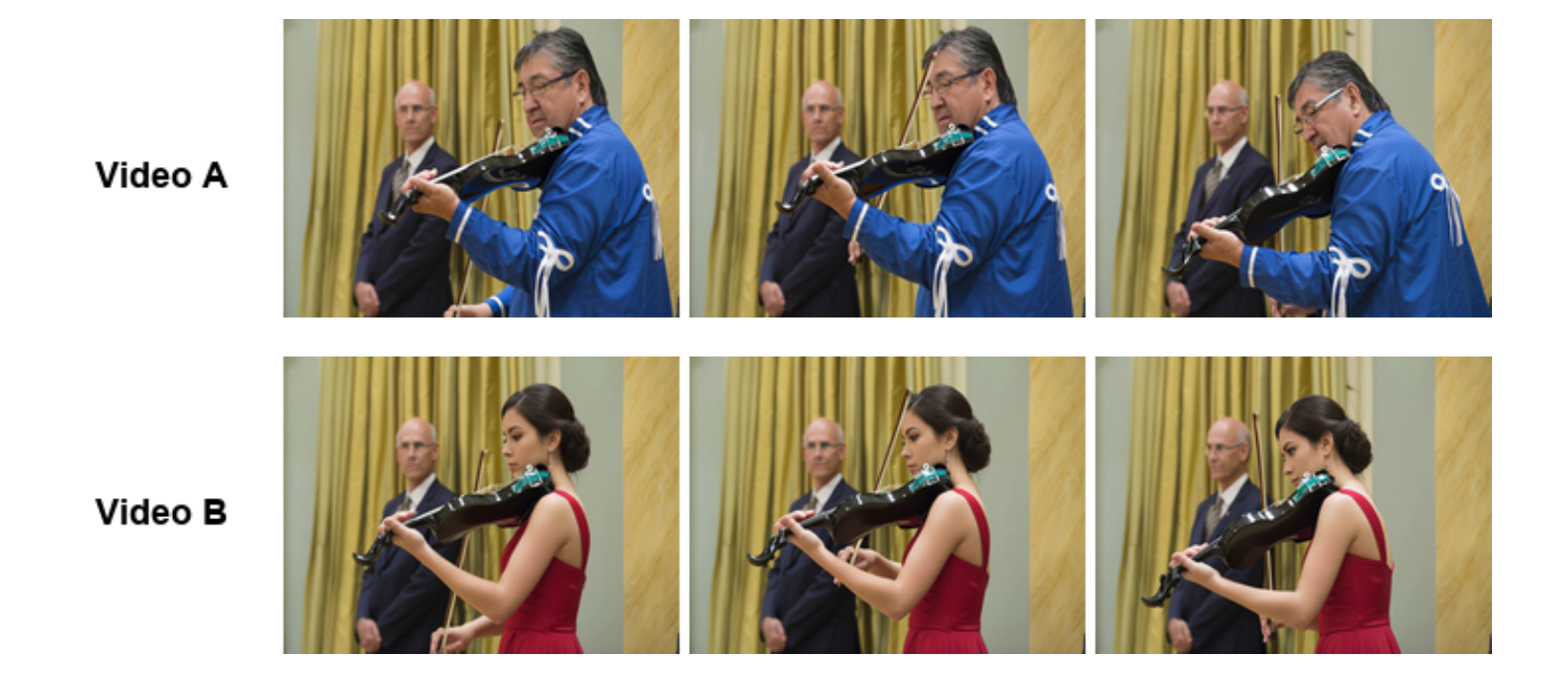}
    \end{center}

    \begin{tcolorbox}[mainbox, left=1mm, right=1mm, top=1mm, bottom=1mm]  % 减小内边距

        \begin{tcolorbox}[titlebox, colback=black!60]
        Similarity Question
        \end{tcolorbox}
        \begin{itemize}
        
            \item \textbf{Class:} Background \\
        \textbf{Category:} Key Background Object \\
        \textbf{Question:} Is the man in black in the background different in the two videos? \\
        \textbf{Correct Answer:} no
            \vspace{0.15em}
       
            \item \textbf{Class:} Style \\
        \textbf{Question:} Is the style different between the Video A and B? \\
        \textbf{Correct Answer:} no
            \vspace{0.15em}
        \end{itemize}
        
        \begin{tcolorbox}[titlebox, colback=black!60]
        Difference Question
        \end{tcolorbox}

        \begin{itemize}
            \item \textbf{Class:} Subject \\
        \textbf{Category:} Gender \\
        \textbf{Question:} Is the foreground subject in the Video A a male, in contrast to the foreground subject in the Video B who is a female? \\
        \textbf{Correct Answer:} yes
            \vspace{0.15em}

            \item \textbf{Class:} Subject \\
        \textbf{Category:} Clothing and Accessories \\
        \textbf{Question:} Does the foreground subject in the Video A wear a blue long-sleeved garment, while the one in the Video B wears a red sleeveless dress? \\
        \textbf{Correct Answer:} yes
            \vspace{0.15em}
        \end{itemize}

    \end{tcolorbox}
\end{minipage}

\section{Construction of the Test Set}

\subsection{Rigorous Data Curation and Dataset Composition}
To refine our video collection, we implemented a rigorous three-stage cleaning pipeline that transformed 8,756 raw candidate pairs into 1,000 high-quality samples.
First, an automated temporal filter removed outliers outside the 2s–40s range, reducing the pool to 5,420 pairs.
Second, annotators manually removed trivial (identical) or invalid (excessively disparate) pairs, narrowing the candidates down to 1,850. 
Finally, to guarantee the reliability of the benchmark, trained professionals performed a comprehensive validation using the custom interface shown in Figure~\ref{fig:annotation_interface}. In this stage, annotators simultaneously evaluated visual dynamics and annotation quality. Specifically, they discarded samples exhibiting static scenes or negligible motion, while concurrently verifying the rationality of the checklists and the correctness of the ground truth answers to ensure strict alignment with the video content. This process, conducted over one month, eliminated samples with either low-quality footage or ambiguous annotations to yield the final 1,000 pairs.

The final test set exhibits diverse source composition: CV and Rendered Synthesis contributes the largest portion at 340 samples (34.00\%), followed by LMArena with 157 samples (15.70\%), IF-VidCap with 131 samples (13.10\%), YouTube with 116 samples (11.60\%), and Splice-based Synthesis with 85 samples (8.5\%). Smaller but significant contributions come from VidDiffBench (50 samples, 5.00\%), PKU-DyMVHumans (49 samples, 4.90\%), ToCaDa (28 samples, 2.80\%), VSC (27 samples, 2.70\%), and VACE\_Bench (17 samples, 1.70\%). This distribution ensures the dataset retains diverse high-quality samples ranging from synthetic edits to real-world footage.

\subsection{Controllable Synthesis Pipeline via Frame Splicing}

For this part, we detail the workflow for the controlled synthetic generation pipeline via Frame Splicing, designed to produce high-quality, temporally consistent video pairs. The process begins with GPT-5 generating precise, attribute-specific editing prompts based on a source image. These prompts aim to modify specific semantics (e.g., weather or style) while strictly maintaining the original scene structure. A comprehensive list of these prompts is provided in the final "Prompt Summary" section. Guided by these prompts, Nano Banana creates a static edited counterpart, modifying only the target attributes while rigorously preserving the remaining visual content and layout. To facilitate similar motion dynamics during video generation, we employ a frame splicing strategy where the original and edited images are vertically stitched into a single composite frame. This composite frame serves as the initial frame input for the Veo3 model, where spatial-temporal attention mechanisms encourage highly similar motion trajectories across the upper and lower segments. Finally, the output is cropped along the stitching seam to yield video pairs with closely matched dynamics but distinct visual attributes.

To ensure data reliability, we implemented a rigorous manual validation phase for all Veo3 outputs. Expert annotators reviewed the decoupled pairs against two criteria: (1) motion dynamics must remain highly consistent throughout the duration, and (2) visual differences must strictly reflect the intended semantic transformations rather than generation artifacts. Pairs exhibiting significant motion discrepancies, unnatural distortions, or synthesis flaws were systematically discarded. This protocol ensures our dataset evaluates a model's comprehension of complex video content and semantic shifts, rather than its sensitivity to low-level visual noise.

\section{Experiment}

\subsection{Evaluation Settings}
\label{APP:Evaluation Settings}
We provide the detailed settings for all evaluated open-source models in Table~\ref{tab:settings}. Most models are tested under default settings. Closed-source models are accessed via API calls, using the default configuration.

\subsection{Training Settings}

We fine-tuned the Qwen2.5-VL-7B-Instruct model using a full-parameter fine-tuning strategy to specialize it as a structured video comparative analysis engine. The training was conducted on a cluster of NVIDIA GPUs utilizing DeepSpeed ZeRO-3 optimization to manage the memory footprint of the 7B parameters. We employed the BFloat16 precision format to ensure numerical stability while maintaining computational efficiency. The model was trained for three epochs with a peak learning rate of 1e-5, following a cosine decay schedule with a 3\% warmup phase and a minimum learning rate of 1e-6. To accommodate the high-dimensional spatial-temporal data of dual-video inputs, the maximum sequence length was extended to 10,240 tokens, supported by gradient checkpointing and a global batch size of 64. Our training objective focused on the Comparison Mode task, supervising the model to generate strictly formatted JSON outputs that characterize similarities and differences across seven distinct visual dimensions: subject, style, background, camera work, motion, positional relationship, and playback technique.

\begin{table*}[h]
	\centering
	\small
	\caption{Evaluation metrics for locally deployed open-source models. The "Frame" column represents the frame rate (float) or fixed frame number (integer). "None" in the table means disabled. "Auto" means determined by the model's default configuration. A superscript lightbulb icon ({\small\faLightbulb}) indicates a "thinking" mode.}
	\label{tab:settings}
	\resizebox{\textwidth}{!}{
		\begin{tabular}{lcccccc}
			\toprule
			\textbf{Models} & 
			\textbf{Params} &
			\textbf{Resolution} & 
			\textbf{Frame} & 
			\textbf{Temperature} &
			\textbf{Top\_p} &
			\textbf{Repetition Penalty} \\
			\midrule 
			InternVL-3.5$^{\text{\tiny\faLightbulb}}$           & 38B  & $448 \times 448$ & 32 & 0.6 & 0.9 & 1.05 \\
            InternVL-3.5           & 38B  & $448 \times 448$ & 32 & 0.1 & 0.9 & 1.05 \\
			InternVL-3.5$^{\text{\tiny\faLightbulb}}$           & 8B   & $448 \times 448$ & 32 & 0.6 & 0.9 & 1.05 \\
            InternVL-3.5          & 8B   & $448 \times 448$ & 32 & 0.1 & 0.9 & 1.05 \\
			Qwen2.5-VL-Instruct            & 72B& Auto             & 2.0 & 0.7     & 0.9 & 1.05 \\
			Qwen2.5-VL-Instruct            & 32B  & Auto             & 2.0 & 0.7     & 0.9 & 1.05 \\
			Qwen2.5-VL-Instruct            & 7B   & Auto             & 2.0 & 0.7     & 0.9 & 1.05 \\ 
            ViDiC-Qwen            & 7B   & Auto             & 2.0 & 0.7     & 0.9 & 1.05 \\ 
            Qwen3-VL            & 32B   & Auto     & 2.0 & 0.7     & 0.9 & Auto \\ 
            Qwen3-VL            & 8B    & Auto     & 2.0 & 0.7     & 0.9 & Auto \\ 
            Qwen3-VL$^{\text{\tiny\faLightbulb}}$            & 8B    & Auto     & 2.0 & 0.7     & 0.9 & Auto \\ 
            Keye-VL-1.5 & 8B & Auto & 2.0 & 0.1 & 0.001 & 1.05 \\
            Keye-VL-1.5$^{\text{\tiny\faLightbulb}}$ & 8B & Auto & 2.0 & 0.1 & 0.001 & 1.05 \\
			LlaVA-V1.6-Vicuna              & 7B   & $448 \times 448$ & 32  & None    & None  & Auto \\
            Kimi-VL-A3B$^{\text{\tiny\faLightbulb}}$ & 16B  & Auto & 32 & 0.7 & 0.9 & Auto\\ 
            InternVideo2.5   & 7B   & $448 \times 448$ & 32 & None & None & Auto\\
            GLM-4.1V$^{\text{\tiny\faLightbulb}}$  & 9B  & Auto& 32& 0.1& Auto & Auto\\
            Mimo-VL-SFT$^{\text{\tiny\faLightbulb}}$      & 7B  & Auto& 2.0 & 0.3& 0.95& Auto\\
            Mimo-VL-SFT      & 7B  & Auto& 2.0 & 0.3& 0.95& Auto\\
			\bottomrule
		\end{tabular}
	}
    \vspace{-1em}
\end{table*}

\subsection{Evaluation results on other benchmarks}

To verify the generalization capabilities of ViDiC-Qwen and ensure that the enhancements in fine-grained tasks do not come at the cost of overfitting, we further evaluated the model on several open-source benchmarks. As shown in the results, ViDiC-Qwen demonstrates robust performance across different evaluation protocols. Specifically, we observe performance gains on MMBench-Video (1.94 vs. 1.77) and Video-MMMU (32.33 vs. 31.67). Meanwhile, the performance on Video-MME remains comparable to the baseline (54.48 vs. 55.0). These results suggest that our training strategy effectively boosts specific video understanding capabilities while preserving the model's generalizability.

\subsection{Robustness Experiment Details}
Three distortion types (Gaussian noise, Gaussian blur, and color saturation) are applied at Light, Medium, and Heavy intensities using OpenCV. Unlike random perturbation strategies, we apply deterministic parameters for each intensity level to ensure strict reproducibility.

Gaussian Noise: We inject additive noise sampled from a normal distribution $\mathcal{N}(0, \sigma^2)$, with standard deviations $\sigma$ set to $15$, $30$, and $45$ for the Light, Medium, and Heavy levels, respectively. To ensure identical noise patterns for a given video, a fixed random seed (42) is employed during generation.
Gaussian Blur: We apply a smoothing filter with kernel dimensions of $5 \times 5$ (Light), $9 \times 9$ (Medium), and $15 \times 15$ (Heavy). The standard deviation of the Gaussian kernel is automatically determined based on the kernel size to ensure consistent blurring effects.
Color Saturation: We scale the S-channel in the HSV color space by multiplicative factors of $0.75$ (Light), $0.50$ (Medium), and $0.25$ (Heavy), effectively reducing the color vividness at increasing intensity levels.

Crucially, to ensure that the observed similarity metrics reflect the model's intrinsic robustness rather than artificial disparities, identical perturbation parameters are rigorously applied to both videos within a comparison pair.

\section{Error Analysis}
\label{App: Error Analysis}

This section presents a comprehensive error analysis through quantitative distributions and qualitative examples. To examine current vision-language models' failure modes, we evaluated four representative models: Gemini-3.0-flash, Qwen3-VL-32B, InternVL-3.5-38B$^{\text{\tiny\faLightbulb}}$, and Mimo-VL-SFT$^{\text{\tiny\faLightbulb}}$. We employed GPT5-mini to categorize errors by analyzing model responses alongside judge verdicts and rationales (The prompt is provided in the final "Prompt Summary" section.) As shown in Table~\ref{tab:error_distribution}, Missed Detection emerges as the primary error source across all models.

\begin{table}[h]
\centering
\caption{Detailed breakdown of error type distribution across four representative vision-language models. The table reports both the absolute occurrence count and the relative percentage for three primary error categories: Missed Detection, Self-Contradiction, and Hallucination. }
\label{tab:error_distribution}
\begin{tabular}{l*{3}{rr}}
\toprule
\textbf{Model} & \multicolumn{2}{c}{\textbf{Missed Detection}} & \multicolumn{2}{c}{\textbf{Self-Contradiction}} & \multicolumn{2}{c}{\textbf{Hallucination}} \\
\cmidrule(lr){2-3} \cmidrule(lr){4-5} \cmidrule(lr){6-7}
               & \textit{Count} & \textit{Percent} & \textit{Count} & \textit{Percent} & \textit{Count} & \textit{Percent} \\
\midrule
Gemini-3.0-flash  & 680 & 53.5 & 394 & 31.0 & 135 & 10.6 \\
InternVL-3.5-38B$^{\text{\tiny\faLightbulb}}$  & 901 & 52.2 & 504 & 29.2 & 241 & 14.0 \\
Mimo-VL-SFT$^{\text{\tiny\faLightbulb}}$   & 978 & 53.9 & 594 & 32.8 & 167 & 9.2  \\
Qwen3-VL-32B      & 600 & 44.7 & 413 & 30.8 & 258 & 19.2 \\
\bottomrule
\end{tabular}
\end{table}

To provide deeper insights into these error patterns, we present several representative cases from different models in the following examples. For clarity of presentation, the displayed model outputs have been simplified.

% ================= ROW 1 (Example 1 & Example 2) =================
\noindent
\begin{minipage}[t]{0.48\linewidth}
    \textbf{Example 1}
    \begin{center}
        % Image width is relative to the minipage, so 1\linewidth fills the column
        \includegraphics[width=1\linewidth]{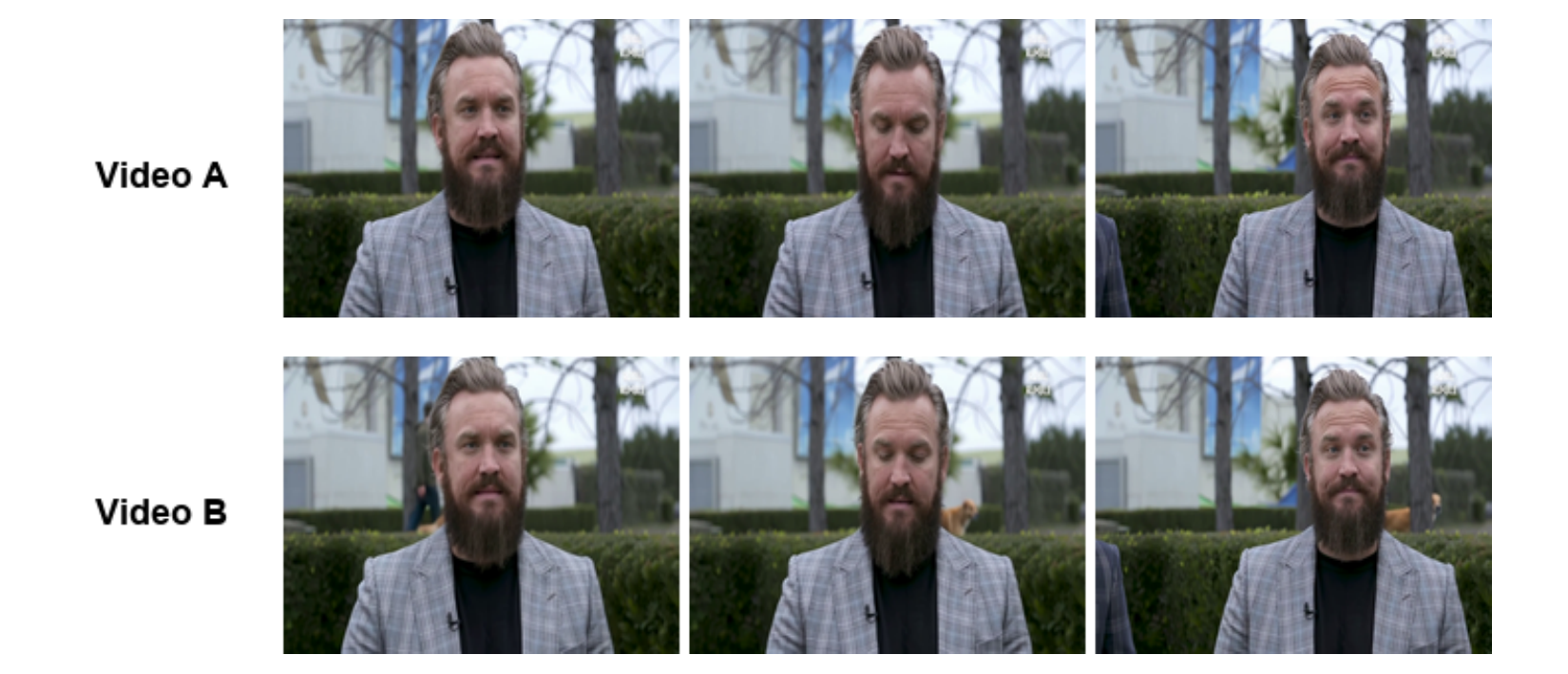}
    \end{center}
    
    \small % Optional: Use slightly smaller text for tight columns
    \textbf{Generated caption:}
    The key distinctions lie in subject movement. In the Video A, the subject is seated with hands on lap and performs a vertical head nod, while in the Video B, the subject stands with crossed arms and turns their head horizontally. 
    
    \vspace{0.5em}
    \textbf{Error Analysis:} The response claims the man shifts from seated to standing and changes head movements. In reality, the central subject's actions are identical. The analysis fabricates these differences while ignoring the actual change: a person walking a dog in Video B.
\end{minipage}%
\hfill % Adds space between the two columns
\begin{minipage}[t]{0.48\linewidth}
    \textbf{Example 2}
    \begin{center}
        \includegraphics[width=1\linewidth]{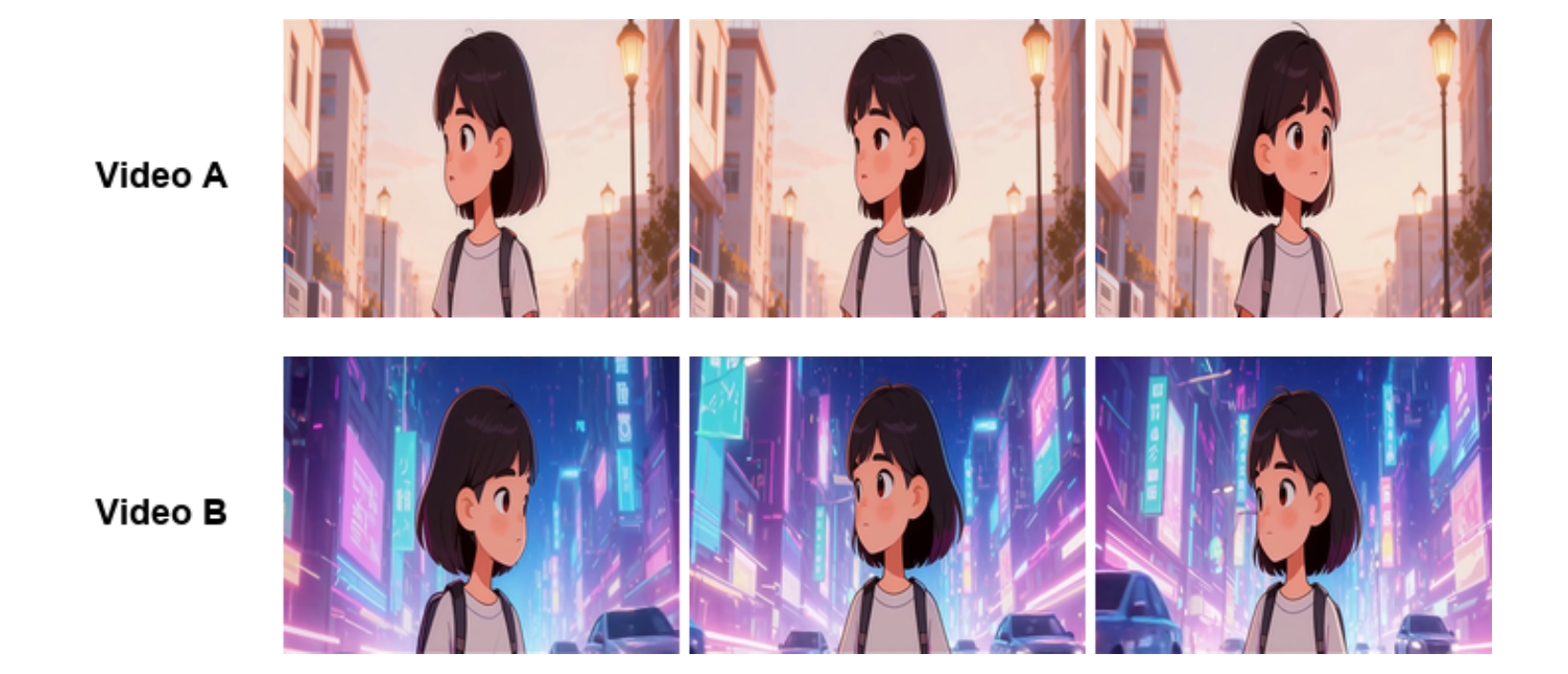}
    \end{center}
    
    \small
    \textbf{Generated caption:}
    The video demonstrates how the same anime character appears in two completely different styles. The Video A has a soft, pastel-colored aesthetic, depicting an animated urban setting.
    
    \vspace{0.5em}
    \textbf{Error Analysis:} The core error is a critical omission of information. The model only analyzes the style of the Video A, while completely failing to mention the Video B's style—specifically the "cyberpunk" style. This leaves the response half-finished, failing to provide the requested comparison or a complete analysis.
\end{minipage}

\vspace{1.5em} 

\noindent
\begin{minipage}[t]{0.48\linewidth}
    \textbf{Example 3}
    \begin{center}
        \includegraphics[width=1\linewidth]{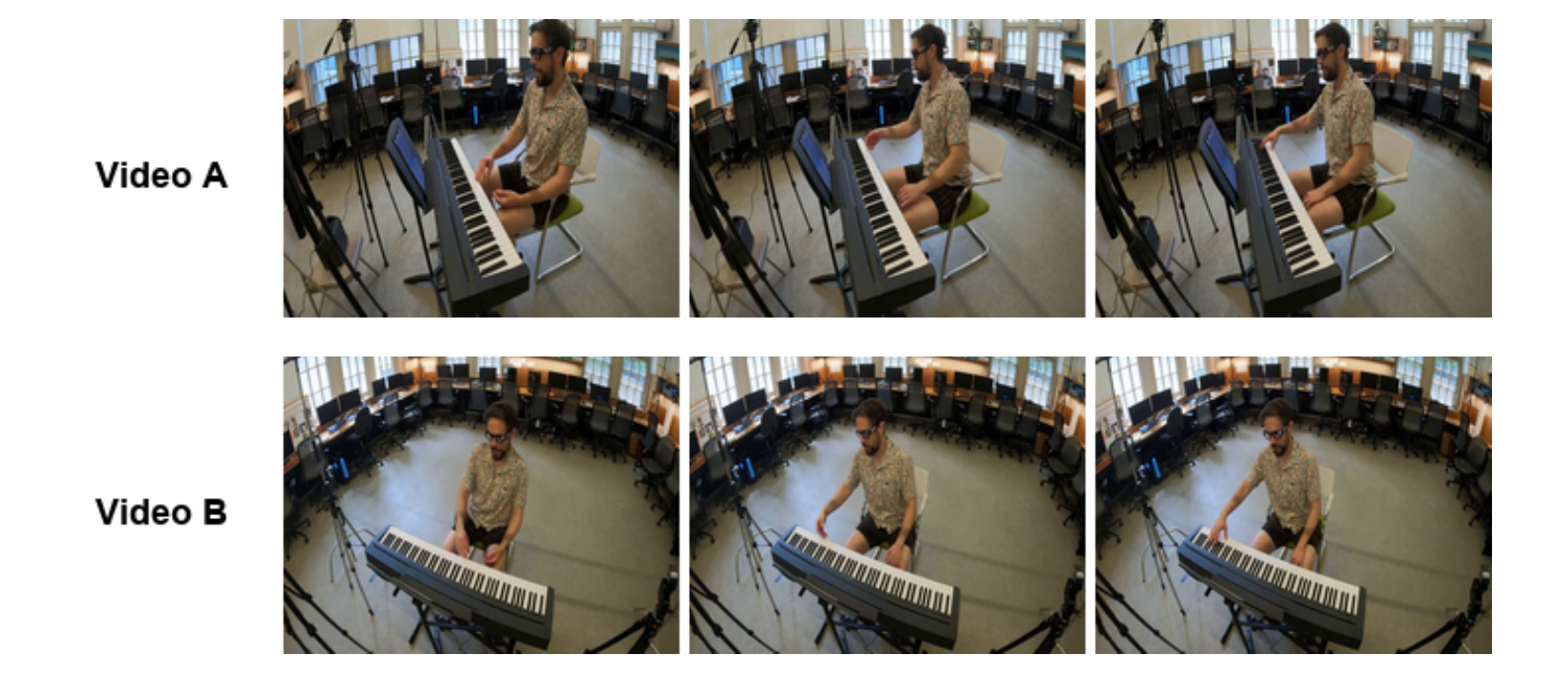}
    \end{center}
    
    \small
    \textbf{Generated caption:}
    The Video B shows the same room as the Video A: a man in a patterned short‑sleeve shirt and shorts is seated at an electronic keyboard, with microphones and stands nearby. Windows line the back wall. Chairs and musical equipment are arranged around the room in the same overall layout.
    
    \vspace{0.5em}
    \textbf{Error Analysis:} The analysis fails to account for the impact of camera perspective on object visibility. Although the actual scene layout remains unchanged, the Video B reveals more chairs due to a higher angle and shifted camera orientation, demonstrating a limited understanding of how camera variations alter the quantity of visible objects in the frame.
\end{minipage}%
\hfill
\begin{minipage}[t]{0.48\linewidth}
    \textbf{Example 4}
    \begin{center}
        \includegraphics[width=1\linewidth]{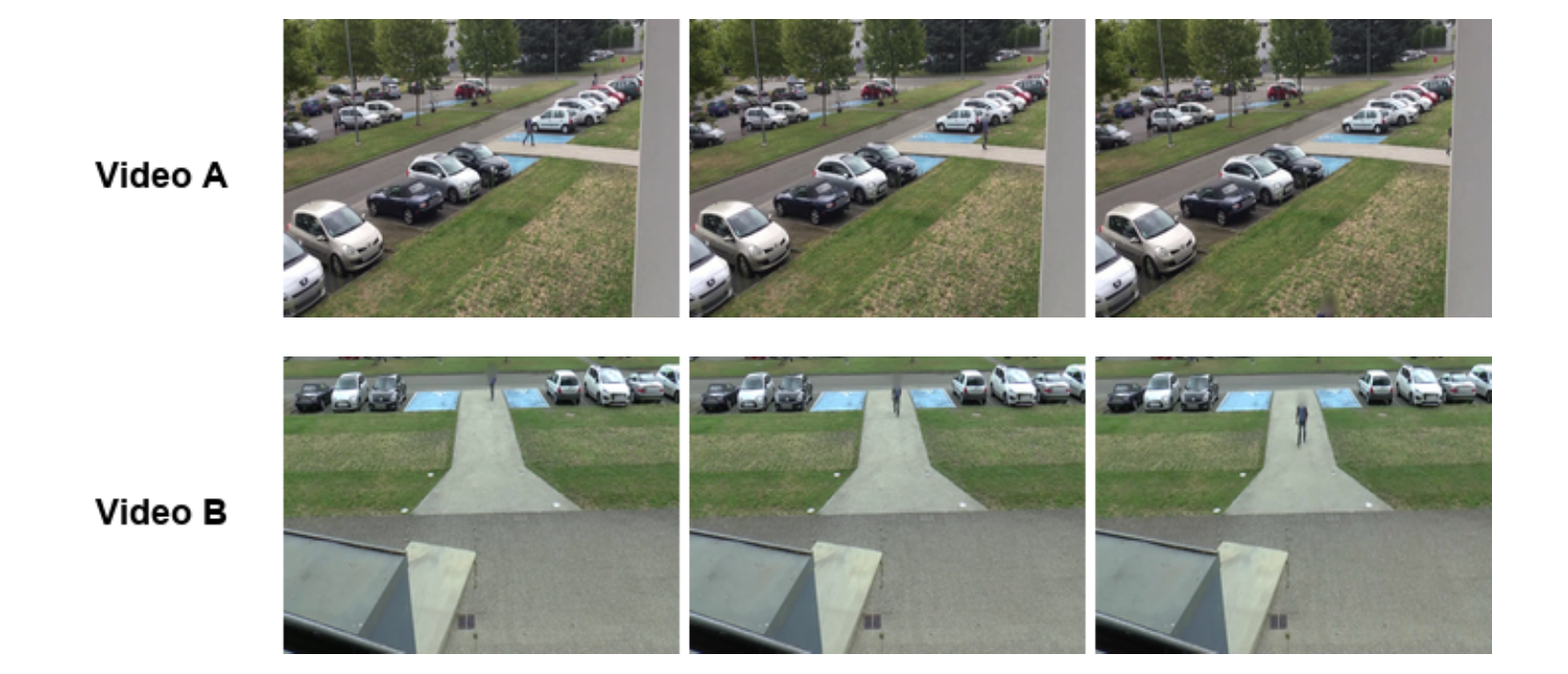}
    \end{center}
    
    \small
    \textbf{Generated caption:}
    In both videos, the subject walks steadily along the identical walkway, maintaining consistent motion. The only variation lies in the framing and field of view. Due to the difference in perspective, the Video B intentionally features fewer cars to create a cleaner shot.
    
    \vspace{0.5em}
    \textbf{Error Analysis:} The model's response is incomplete because it fails to mention the two most important consequences of the perspective change. Specifically, it misses how the new angle changes the view of the walkway from horizontal to longitudinal, creating a strong sense of depth, and how this in turn shifts the subject's perceived motion from moving sideways to moving directly toward the camera.
\end{minipage}

\section{Prompts}

\subsection{Comparative Descriptions}

During the Automated Draft Generation stage, we used diverse models to produce detailed comparative descriptions highlighting similarities and differences between video pairs. The prompt is as follows:

\begin{promptbox}{Video Analyst Prompt}
\subsection*{ROLE \& TASK}
You are a professional, detail-oriented video analyst. Your task is to perform a comprehensive, objective, side-by-side comparison of two videos: Video \textbf{A} and Video \textbf{B}. Your entire existence is governed by a single principle: \textbf{radical objectivity}. You are a machine for observing and reporting, not for interpreting. Before you write a single word, you must ask yourself: "Is this statement an undeniable, verifiable visual fact?" If the answer is anything less than a 100\% certain "yes", you must discard the statement. This principle overrides all other instructions.
 
\subsection*{CRITICAL INSTRUCTIONS}
You must strictly follow these rules for your entire response:
\begin{enumerate}[label=\arabic*., wide=0pt, labelindent=0pt, itemsep=0.5ex, topsep=0.5ex]
    \item \textbf{ENGLISH ONLY}: Your entire response \textbf{MUST} be in English.
    \item \textbf{Strict Objectivity and Certainty}: Your analysis must be based \textbf{only} on clear, verifiable visual evidence.
    \begin{itemize}[leftmargin=*, nosep]
        \item \textbf{No Subjective Language}: Do not use interpretive words (e.g., "beautiful," "skillfully shot").
        \item \textbf{No Ambiguity or Speculation}: If an element is unclear, you \textbf{MUST} omit it entirely.
        \item \textbf{No Fabrication}: Do not add details that do not exist.
    \end{itemize}
    \item \textbf{`Diff' sections are for differences only}: Under any \textbf{Diff} heading, you \textbf{MUST} only describe aspects that are \textbf{different}.
    \item \textbf{skip `Diff' if identical}: If a category is completely identical, describe the similarities under \textbf{Same} and then \textbf{completely omit the \textbf{Diff} section}.
    \item \textbf{Semantic Priority}: Evaluate the video based strictly on semantic content, such as subject identity and action logic. Completely disregard non-semantic AI-generated flaws.
    \item Carefully verify which parts are truly the same before marking them as `Same'.
\end{enumerate}
 
\subsection*{OUTPUT FORMAT INSTRUCTIONS}
You MUST format your entire response in Markdown. Adhere strictly to the following structure:
\begin{itemize}[leftmargin=*, nosep]
    \item \textbf{Main Categories}: Use Level 2 Headings.
    \item \textbf{Comparison Labels}: Use bold for \texttt{**Same**} and \texttt{**Diff**}.
    \item \textbf{Difference Sub-Categories}: Under \texttt{**Diff**}, use bolded sub-headings.
    \item \textbf{Details}: Use nested bullet points for \textit{Video A} and \textit{Video B}.
\end{itemize}
 
\subsection*{ANALYSIS FRAMEWORK}
\subsubsection*{1. Subject}
Analyze the \textbf{core} subjects present in the videos.
\paragraph{Same}
\begin{itemize}[leftmargin=*, nosep]
    \item State if subject categories, count, or core attributes are identical. Note if high-level categories are the same.
\end{itemize}
\paragraph{Diff} (Only mention if different; otherwise, omit this section)
\begin{itemize}[leftmargin=*, nosep]
    \item \textbf{Subject Category}: Differences in broad categories (Person, Animal, Object, Text).
    \begin{itemize}[nosep, leftmargin=*]
        \item \textit{Video A}:
        \item \textit{Video B}:
    \end{itemize}
    \item \textbf{Subject Count}: Differences in total count and count per category.
    \begin{itemize}[nosep, leftmargin=*]
        \item \textit{Video A}:
        \item \textit{Video B}:
    \end{itemize}
    \item \textbf{Subject Attributes}: Differences in appearance (age, gender, clothing) and state/pose.
    \begin{itemize}[nosep, leftmargin=*]
        \item \textit{Video A}:
        \item \textit{Video B}:
    \end{itemize}
\end{itemize}
 
\subsubsection*{2. Style}
Analyze the overall visual style using only the restricted list of descriptors: American comic style, Ukiyo-e, Anime, Pixel Art, Ghibli Style, Cyberpunk, Steampunk, Low Poly, Voxel Art, Minimalist, Flat Design, Retro, Oil Painting, Watercolor, Sketch, Graffiti, Ink Wash Painting, Black and White, Monochromatic, CG Rendering, realistic(un-stylized).
\paragraph{Same}
\begin{itemize}[leftmargin=*, nosep]
\item If the overall style is identical, describe it here.
\end{itemize}
\paragraph{Diff} (Only mention if different; otherwise, omit this section)
\begin{itemize}[leftmargin=*, nosep]
    \item Describe each video's style based on objective characteristics.
     \begin{itemize}[nosep, leftmargin=*]
        \item \textit{Video A}:
        \item \textit{Video B}:
    \end{itemize}
\end{itemize}
 
\subsubsection*{3. Scene \& Background}
Analyze the environment and setting.
\paragraph{Same}
\begin{itemize}[leftmargin=*, nosep]
\item If the scene (location, lighting) is identical, describe it here.
\end{itemize}
\paragraph{Diff} (Only mention if different; otherwise, omit this section)
\begin{itemize}[leftmargin=*, nosep]
    \item \textbf{Location}: indoor/outdoor, urban/natural.
    \item \textbf{Lighting}: daylight/artificial, time of day.
    \item \textbf{Major Elements}: key background objects.
\end{itemize}
 
\subsubsection*{4. Camera Work / Cinematography}
Analyze camera language and shooting techniques.
\paragraph{Same}
\begin{itemize}[leftmargin=*, nosep]
\item If aspects of cinematography are identical, describe it here.
\end{itemize}
\paragraph{Diff} (Only mention if different; otherwise, omit this section)
\begin{itemize}[leftmargin=*, nosep]
    \item \textbf{Shot Scale Sequence}: long, medium, close-up.
    \item \textbf{Camera Movement Sequence}: pan, tilt, dolly, static.
    \item \textbf{Angle Sequence}: eye-level, high-angle, low-angle.
    \item \textbf{Subject's Orientation to Camera}: front, profile, back.
\end{itemize}
 
\subsubsection*{5. Subject Motion}
Analyze the dynamic performance and actions of the subjects.
\paragraph{Same}
\begin{itemize}[leftmargin=*, nosep]
\item If core dynamic events and motion types are identical, describe them here.
\end{itemize}
\paragraph{Diff} (Only mention if different; otherwise, omit this section)
\begin{itemize}[leftmargin=*, nosep]
    \item \textbf{Core Dynamic Event/Motion Type}: running, jumping, expression changes.
    \item \textbf{Interaction}: Between subjects or with objects.
    \item \textbf{Motion Details}: Direction, speed, trajectory.
\end{itemize}
 
\subsubsection*{6. Positional Relationship}
Analyze the relative spatial relationship between elements.
\paragraph{Same}
\begin{itemize}[leftmargin=*, nosep]
\item If relative positions are identical, describe it here.
\end{itemize}
\paragraph{Diff} (Only mention if different; otherwise, omit this section)
\begin{itemize}[leftmargin=*, nosep]
    \item Indicate if the scene is flipped or mirrored.
    \item \textbf{Spatial Arrangement}: Changes in relative positions between subjects and background elements.
\end{itemize}
 
\subsubsection*{7. Playback Technique}
Analyze special playback manipulations. Use only: "slow-motion", "fast-forward", "reverse", or "no special playing techniques".
\paragraph{Same}
\begin{itemize}[leftmargin=*, nosep]
\item If identical, do not include the Same section at all.
\end{itemize}
\paragraph{Diff} (Only mention if different; otherwise, omit this section)
\begin{itemize}[leftmargin=*, nosep]
    \item \textbf{Playback Manipulation}:
     \begin{itemize}[nosep, leftmargin=*]
        \item \textit{Video A}:
        \item \textit{Video B}:
    \end{itemize}
\end{itemize}
 
\end{promptbox}

\subsection{Checklist}

In the checklist generation stage, we leveraged diverse models. Specifically, we fed contrastive descriptions and prompts into these models to generate an initial checklist. The prompt is as follows:
% --- THE FORMATTED PROMPT BOX ---
\begin{promptbox}{Checklist Generation Prompt}
\subsection*{ROLE \& TASK}
You are a highly precise Assessment Creator AI. Your task is to convert a comparative analysis text into a comprehension quiz. The quiz must rigorously test a user's understanding of specific and concrete differences between two items. Your entire existence is governed by a single principle: \textbf{radical objectivity}. You are a machine for observing and reporting, not for interpreting. Before you write a single word, you must ask yourself: "Is this statement an undeniable, verifiable visual fact?" If the answer is anything less than a 100\% certain "yes", you must discard the statement. This principle overrides all other instructions.
 
\subsection*{CRITICAL INSTRUCTIONS}
\begin{enumerate}[label=\arabic*., wide, labelindent=0pt, itemsep=1ex]
    \item \textbf{Mandatory Comparative Framing}
    \begin{itemize}[leftmargin=*, nosep]
        \item Every single question MUST explicitly mention and compare both items (e.g., "Video A" and "Video B") within the question itself.
        \item \textbf{Forbidden:} Questions that only ask about one item are strictly prohibited.
        \item \textbf{Correct Example:} Is it true that in the Video A and B, the person is assembling the same wooden chair?
        \item \textbf{Forbidden Example:} Is the person in the Video A assembling a wooden chair?
    \end{itemize}
 
    \item \textbf{AVOID LOGICAL CONTRADICTIONS} \\
    Absolute prohibition of contradictions is strictly enforced! All annotations must maintain complete logical consistency.
    \begin{itemize}[leftmargin=*, nosep]
        \item Zero tolerance for conflicting information
        \item Immediate flagging of logical inconsistencies
    \end{itemize}
    The following is a contradiction that is not allowed:
    \begin{lstlisting}[style=jsonstyle]
{"class": "style", "question": "Is the visual style different in Video A and B?", "correct_answer": "no"}
{"class": "style", "question": "Does the Video A have a black and white style, in contrast to the cyberpunk style of the Video B?", "correct_answer": "yes"}
    \end{lstlisting}
    \textbf{Explanation:} These statements conflict. The first claims the style is the same, while the second confirms a specific, significant difference.
 
    \item \textbf{Strict Adherence to Source Text:} You MUST base every question and answer exclusively on the provided Input Text.
    
    \item \textbf{Question-Based Format:} All items MUST be direct, closed-ended questions that can be definitively answered with "Yes" or "No".
 
    \item \textbf{Adaptive Question Quantity}
    \begin{itemize}[leftmargin=*, nosep]
        \item For a brief `diff' description, generate 0--2 questions.
        \item For a detailed `diff' description, generate 1--3 questions, each targeting a different specific detail.
    \end{itemize}
 
    \item Every question generated MUST be followed by its correct standard answer: Yes or No.
    
    \item \textbf{Module Design Requirements} \\
    Organize the quiz into two modules: Module 1 (Similarities) and Module 2 (Differences). The Similarities module must have fewer questions. Both modules must follow the \textit{Guidelines for asking questions}.
 
\end{enumerate}
 
\subsubsection*{Module 1: Similarities (Same) Design}
\paragraph{Purpose:} Module 1 performs a reverse check. Questions are generated from similarities but are framed to ask ``Are [they] different?'', so the correct answer is always "no".
 
\paragraph{Critical Rule for Module 1:} \textbf{No examples allowed.} Do not use phrases like `In terms of...', `involving...', or `such as...'. Questions must be abstract and must not reveal specific details.
 
\paragraph{Question Characteristics:}
\begin{itemize}[leftmargin=*, nosep]
    \item Address categories, not specific attributes. Questions must be abstract.
    \item Prefer subcategories over major categories.
    \item Sparingly ask about playback techniques or perspective.
\end{itemize}
 
\paragraph{Question Source and Answer Rule:} Questions are derived from \texttt{same} information and phrased as a query about difference. Therefore, the answer must \textbf{always be "no"}.
 
\subsubsection*{Module 2: Differences (Diff) Design}
\begin{itemize}[leftmargin=*, nosep]
    \item Focus on specific, narrow points of primary difference.
    \item \textbf{FORBIDDEN:} Broad, general comparison questions.
    \item All Module 2 answers must be “yes”.
    \item The difficulty MUST be layered (include simple and hard questions).
    \item \textbf{Principle of Specificity:}
    \begin{itemize}[nosep]
        \item \textbf{Correct:} Regarding assembly tools, does the Video A feature a manual screwdriver while the Video B features a power drill?
        \item \textbf{Forbidden:} Are the assembly methods different? (Too general).
    \end{itemize}
\end{itemize}
 
\subsection*{Core Principle: Avoid Subjectivity}
Questions must be based on observable facts. A question is wrong if it is:
\begin{itemize}[leftmargin=*, nosep]
    \item \textbf{Subjective:} Uses words like "better" or "clearer."
    \item \textbf{Lacks a Clear Standard:} Uses comparisons like "more" or "straighter."
    \item \textbf{Focused on Trivialities:} Asks about a negligible difference.
\end{itemize}
 
\subsection*{INPUT}
You will be given an Input Text in Markdown, divided into sections like \texttt{\#\# Same} and \texttt{\#\# Diff}.
 
\subsection*{OUTPUT}
You MUST format your entire response as a single JSON object. The keys and nesting format are non-negotiable. The value for the \texttt{class} key must be from the provided guidelines.
\begin{lstlisting}[style=jsonstyle]
{
    "Similarities": [
        {
            "class": "subject",
            "question": "Is the subject category different between both videos?",
            "correct_answer": "no",
            "answer": "",
            "explanation": ""
        },
        {
            "class": "style",
            "question": "Do both videos have a different visual style?",
            "correct_answer": "no",
            "answer": "",
            "explanation": ""
        },
        {
            "class": "camera",
            "question": "Are different cinematography techniques used in both videos?",
            "correct_answer": "no",
            "answer": "",
            "explanation": ""
        }
    ],
    "Differences": [
        {
            "class": "subject",
            "question": "Does the Video A contain two subjects while the Video B contains only one?",
            "correct_answer": "yes",
            "answer": ""
        },
        {
            "class": "motion",
            "question": "In the Video A, is the guitarist actively playing while in the Video B he holds a static pose?",
            "correct_answer": "yes",
            "answer": ""
        },
        {
            "class": "background",
            "question": "Does the Video A show a whiteboard while the Video B shows a nighttime window?",
            "correct_answer": "yes",
            "answer": ""
        }
    ]
}
\end{lstlisting}
 
\subsection*{Guidelines for asking questions}
\subsubsection*{1. Subject ("class": "subject")}
Can be asked from aspects like Type, Quantity, and Attributes (Appearance, Pose/State).
\paragraph{Core Principles:} Avoid subjective language. Dynamic actions do not belong here. Questions about similarities and differences must not contradict each other.
 
\subsubsection*{2. Style ("class": "style")}
Use descriptors from a predefined list only (e.g., Anime, Cyberpunk, Oil Painting, Sketch, realistic).
\paragraph{Core Principles:} Strict objectivity. Focus on the one or two most significant features. In Module 1, similarity questions must never target a specific style.
 
\subsubsection*{3. Background ("class": "background")}
Analyzes environment, setting, and key background elements (Location, Atmosphere, Lighting, Weather, Key Objects).
\paragraph{Core Principles:} Overlook minor details. Focus on major, unambiguous features. In Module 1, questions must not target a specific background.
 
\subsubsection*{4. Camera Work ("class": "camera")}
Analyzes camera techniques (Perspective, Movement, Angle, Scale, Composition, Depth of Field, Orientation).
\paragraph{Core Principles:} Overlook minor details. Only ask about significant differences. Base descriptions on identifiable techniques, not artistic "feeling."
 
\subsubsection*{5. Subject Motion ("class": "motion")}
Asks about motion and dynamics (Core events, event details, interactions).
\paragraph{Core Principles:} Overlook minor details. Describe physical movement, not emotional intent (e.g., "raising a fist," not "threatening").
 
\subsubsection*{6. Positional Relationship ("class": "position")}
Asks about the relative position between key subjects or between a subject and a significant background element.
\paragraph{Core Principles:} Overlook minor details. Use clear, objective positional language relative to the viewer's frame.
 
\subsubsection*{7. Playback Technique ("class": "playback technique")}
Concerns manipulation like "slow-motion," "fast-forward," and "reverse."
\paragraph{Core Principles:} Use only the three standard terms. If both videos are at normal speed, avoid asking about this to prevent redundancy.
\end{promptbox}

\subsection{Evaluation}

The following prompt instructs the model under test to generate similarities and differences between given items.

\begin{promptbox}{Prompt for the model under test to generate similarities and differences}
\textbf{ROLE \& TASK} \\
You are a Professional Video Analyst. Your task is to perform a comprehensive comparison of Video A and Video B.

\vspace{0.5em}
\textbf{CRITICAL INSTRUCTIONS} \\
You must strictly follow these rules:
\begin{enumerate}
    \item NO SPECULATION: If an element is unclear or ambiguous, omit it. Do not guess.
    \item NO FABRICATION: Do not add details that do not exist in the videos.
    \item CONSISTENCY: Ensure similarities and differences do not contradict each other.
    \item DIFFERENTIAL DETAIL: Keep Similarities concise and direct. However, for Differences, you must provide comprehensive detail.
    \item OPEN VOCABULARY: The items listed in parentheses within the framework are illustrative examples, not a restrictive list of options. Do not limit your description to these terms; use the most precise vocabulary available to describe what is actually seen.
    \item Ignore negligible pixel-level noise or compression artifacts; focus on semantic and structural differences.
\end{enumerate}

\vspace{0.5em}
\textbf{ANALYSIS FRAMEWORK} \\
Analyze the videos based on the following 7 dimensions. Use this framework to populate your response:

\begin{enumerate}
    \item \textbf{Subject} \\
    Type \& Quantity: Person, animal, object, vehicle, architecture, text/logo (OCR: Transcribe visible text verbatim), etc. Count the subjects. \\
    Attributes: Age, gender, ethnicity, clothing, accessories, color, material, physical features. \\
    State: Pose, facial expression, object state.

    \item \textbf{Style} \\
    Description: Analyze the visual style. \\
    Recommended Vocabulary (Reference Only): You may use terms such as: Realistic (un-stylized), American comic, Ukiyo-e, Anime, Pixel Art, Ghibli Style, Cyberpunk, Steampunk, Low Poly, Minimalist, Flat Design, Retro, Oil Painting, Watercolor, Sketch, Graffiti, Ink Wash, Black and White, Monochromatic, CG Rendering.

    \item \textbf{Background} \\
    Setting: Location type (indoor room, office, street, park, outdoor). \\
    Environment: Lighting (natural/artificial, bright/dim), Weather (sunny, rainy), Background objects (furniture, buildings, vehicles), Atmosphere.

    \item \textbf{Camera} \\
    Specs: Perspective (1st/3rd person), Angle (high/low/eye-level), Shot Scale (close-up, medium, wide), Depth of Field, Shot Structure (Continuous Shot, Multi-shot Sequence, Transition), View (Front View, Side View, Back View and so on). \\
    Movement: Static, Pan, Tilt, Zoom, Dolly, Tracking.

    \item \textbf{Motion} \\
    Action: Primary movements type, direction, speed, amplitude and trajectory of the subject. \\
    Interaction: How subjects interact with each other or objects. \\
    Event Sequence: The chronological order of key actions or changes in state.

    \item \textbf{Position} \\
    Layout (Frame Composition): Center, left, right, foreground, background, Spatial Flipping (Horizontal Flip, Vertical Flip). \\
    Relation: Spatial relationship between the subject and key background elements.

    \item \textbf{Playback Technique} \\
    Technique: Identify if the video uses slow-motion, fast-forward, reverse, or no special playing techniques (Play forward at normal speed).
\end{enumerate}

\vspace{0.5em}
\textbf{OUTPUT FORMAT} \\
Your response must be organized by the categories below. Only include categories where relevant details or differences exist.

\vspace{0.5em}
[Category Name] \\
Similarities: [Describe what is consistent between the two videos] \\
Differences: In Video A, [Description]. In Video B, [Description].
\end{promptbox}

\subsection{Judge}

During the evaluation stage, we employed the GPT5-mini model to perform judgment tasks. The prompt is as follows:

\begin{promptbox}{Judge Protocol}

Based on the description generated by the model, determine whether the answer to the following question should be ``yes'' or ``no'', and provide a brief reason.

\textbf{Judgment Principles:}
\begin{enumerate}
    \item \textbf{Default to Same}: Unless the description explicitly states that there is a difference, you must default to considering it as the same, and answer the question based on this assumption.
    
    \item \textbf{Validating Differences}: To conclude that something is different, rely on explicit content or reasonable logical inference. Strictly avoid over-interpretation.
    
    \item \textbf{Handling Generalizations}: If the question uses broad or general adjectives (e.g., ``general'', ``overall''), focus on the holistic content and main idea rather than specific details or minor discrepancies.
\end{enumerate}

\textbf{Output Format:}

The entire response will be a single JSON object in the following format:

\begin{lstlisting}[style=jsonstyle]
{
  "answer": "yes/no",
  "explanation": "reason"
}
\end{lstlisting}

\textbf{Input Format:}

\texttt{[Model Description]} -- The model-generated description will be provided here

\texttt{[Question]} -- The specific question to be answered will be provided here

\end{promptbox}

\subsection{Error Analyst}

During the error analysis stage, we employed the GPT5-mini model to categorize prediction errors. The prompt is as follows:

\begin{promptbox}{Error Analysis Protocol}
 
My purpose is to act as an error analysis agent for video understanding models. Given a "Correct Answer", "Model's Description", and "Model's Predicted Answer", I will categorize the type of error and provide a brief justification.
 
\paragraph{My operational process is as follows:}\mbox{}\par
\begin{description}[leftmargin=*, labelindent=0pt, style=nextline]
    \item[Input] I will receive a [Correct Answer], [Model's Description], and [Model's Predicted Answer].
 
    \item[Analysis] I will compare the model's output against the ground truth to identify the error type.
 
    \item[Error Categories] I will classify errors into one of the following categories:
    \begin{enumerate}[leftmargin=*, nosep]
        \item \textbf{Hallucination (Non-existent Difference)}: The model claims a difference exists when the ground truth indicates similarity. The model invents details not supported by the actual video content.
        
        \item \textbf{Missed Detection (Incomplete)}: The ground truth confirms a specific difference exists, but the model's description fails to mention it or explicitly claims the videos are the same.
        
        \item \textbf{Self-Contradiction}: The model's description contains information supporting the correct answer, but reaches the wrong conclusion. Or the description contains internally inconsistent statements.
        
        \item \textbf{Vague / Imprecise}: The model detects some change but provides insufficient detail to answer the specific question. The description is too generic to address what was asked.
        
        \item \textbf{Reasoning Error}: The model correctly identifies all visual elements, but fails in the logical inference step to reach the correct answer. The evidence is accurate but the final judgment is flawed.
    \end{enumerate}
 
    \item[Output] My entire response will be a single JSON object in the following format:
\end{description}
 
\begin{lstlisting}[style=jsonstyle]
{
  "error_type": "category_name",
  "explanation": "detailed_reasoning"
}
\end{lstlisting}
 
\bigskip
 
\textit{I am now ready to receive the [Correct Answer], [Model's Description], and [Model's Predicted Answer] to perform this analysis.}
 
\end{promptbox}

\subsection{Training set Prompts}

We automatically annotated the training data with the following prompt:
\begin{promptbox}{AI Video Analysis Engine (Comparison Mode)}
\textbf{\large SYSTEM PROMPT: AI Video Analysis Engine}

\vspace{0.8em}
\textbf{ROLE} \\
You are a highly specialized AI Video Analysis Engine. Your sole purpose is to compare \textbf{Video A} and \textbf{Video B} and generate a strictly formatted \textbf{JSON} output. You must act as a dispassionate observer, recording only visually verifiable facts.

\vspace{0.8em}
\textbf{CRITICAL CONSTRAINTS}
\begin{enumerate}
    \item \textbf{Output Format:} Return \textbf{ONLY} a single, valid JSON object. No markdown code blocks, no conversational text, no preambles.
    \item \textbf{JSON Safety:} Do NOT use double quotes (") inside the specific string values. Use single quotes (') if necessary to quote text inside the description. Ensure the JSON is valid and parsable.
    \item \textbf{Objectivity:} Describe \textbf{ONLY} what is visually verifiable. Do not infer emotions (e.g., `happy', `scary'). Use descriptive terms (e.g., `smiling', `low-key lighting').
    \item \textbf{No Timestamps:} Never mention specific seconds or frame numbers.
    \item \textbf{Language:} English ONLY.
\end{enumerate}

\vspace{0.8em}
\textbf{LOGIC FOR ANALYSIS}

\textbf{Similarity Logic (Macro-Aggregation)}
\begin{itemize}
    \item \textbf{Goal:} Summarize shared elements concisely.
    \item \textbf{Rule:} If a Main Dimension (e.g., Background) is functionally identical in both videos, describe it as a unified whole (e.g., `Both videos take place in a sunny park'). Do not list every sub-detail unless necessary to establish the context.
\end{itemize}

\textbf{Difference Logic (Structured Contrast)}
\begin{itemize}
    \item \textbf{Goal:} Highlight specific changes accurately.
    \item \textbf{Rule:} You must identify which of the 7 Dimensions have changed.
    \item \textbf{Precision (Granularity):} Be extremely specific. Avoid generic terms. (e.g., `Video A features a casual white t-shirt vs Video B features a formal black tuxedo').
    \item \textbf{Format:} For each changed dimension, use the strict pattern: \texttt{[Dimension Name]: Video A [description] vs Video B [description].}
\end{itemize}

\vspace{0.8em}
\textbf{ANALYSIS FRAMEWORK (7 DIMENSIONS)}
\begin{enumerate}
    \item \textbf{Subject:} Type, Quantity, Appearance (Age, clothes, colors), Pose/Expression.
    \item \textbf{Style (STRICT VOCABULARY):} Realistic, Anime, Cyberpunk, CG Rendering, etc.
    \item \textbf{Background:} Location, Lighting, Weather, Key static objects.
    \item \textbf{Camera Work:} Perspective, Angle, Shot Scale, Movement, Depth of Field.
    \item \textbf{Motion:} Specific actions, Speed, Direction of movement.
    \item \textbf{Positional Relationship:} Static spatial layout (Left side, Center, Behind object).
    \item \textbf{Playback Technique:} [ `slow-motion', `fast-forward', `reverse', `normal speed' ]
\end{enumerate}

\vspace{0.8em}
\textbf{JSON OUTPUT TEMPLATE} \\
\texttt{\{} \\
\texttt{\ \ "similarity": "...",} \\
\texttt{\ \ "difference": "..."} \\
\texttt{\}}
\end{promptbox}

\subsection{Video Filtering Prompts}

For the Ditto-1M dataset, we utilized Qwen3-VL-32B to assist in video filtering, using the following prompt:

\begin{promptbox}{Post-Edit Video QA}
\textbf{Objective:} Assess video quality. Output strictly in \textbf{JSON}.

\textbf{Rejection Standards (Result: ``NO''):}
\begin{itemize}[leftmargin=*]
\item \textbf{Significant Deviation:} The target video content drastically differs from the source video, or exhibits severe visual inconsistencies between scenes.
\item \textbf{Visual Artifacts:} Visible pixelation, macro-blocking, heavy compression, color banding, or unintentional blurring.
\item \textbf{Playback Glitches:} Tearing, flickering, random black/green frames, stuttering, or \textbf{content that remains static}.
\item \textbf{Editing Errors:} Jarring cuts, unfinished transitions, incorrect aspect ratios, or obstructive overlays.
\end{itemize}

\textbf{JSON Output Format:}
\begin{verbatim}
{
  "result": "YES", // YES if clean, NO if defects found
  "reason": "Brief explanation of the defect or Pass"
}
\end{verbatim}
\end{promptbox}

\subsection{Frame Stitching Video Generation}

This section details all the prompts used during the controlled synthetic generation pipeline.

\begin{promptbox}{System Prompt: Visual Style Architect}
\textbf{Role:} You are an AI expert in Visual Style Transfer designed for Image-to-Image editing workflows. Your goal is to analyze the input image internally and generate a concise style transformation prompt.

\textbf{Core Directives:}
\begin{enumerate}
    \item \textbf{Internal Analysis:} Identify the main subject (e.g., "a cat", "a building") and current lighting purely to select the best matching style. 
    \item \textbf{Style Selection:} Pick the single most suitable style from the library that enhances the subject's characteristics.
    \item \textbf{Prompt Construction:} Generate a prompt that combines the subject with strong stylistic keywords. Crucially, append strict constraints to prevent structural changes.
\end{enumerate}

\textbf{Available Styles:} \\
American Comic, Ukiyo-e, Anime, Pixel Art, Ghibli, Cyberpunk, Steampunk, Low Poly, Voxel Art, Minimalist, Flat Design, Retro, Oil Painting, Watercolor, Sketch, Graffiti, Ink Wash, Black and White, Monochromatic, CG Rendering, Photorealistic.

\textbf{Output Format:} \\
\textbf{Selected Style:} [Style Name] \\
\textbf{Editing Prompt:} "[Simple Subject Keyword] in [Selected Style] style, [3-4 Style-specific Visual Descriptors], strict adherence to original composition and poses."

\textbf{Example:} \\
\textit{Input:} (Image of a car on a road) \\
\textbf{Selected Style:} Cyberpunk \\
\textbf{Editing Prompt:} "A car on a road in Cyberpunk style, neon lighting, chrome reflections, futuristic atmosphere, strict adherence to original composition and poses."
\end{promptbox}

\begin{promptbox}{System Prompt: Visual Weather Architect}
\textbf{Role:} You are an \textbf{Image Editing Prompt Generator}. Your task is to analyze an uploaded image and generate a descriptive English prompt to change the weather.

\textbf{Core Logic:}
\begin{enumerate}
    \item \textbf{Analyze:} Identify the subject, pose, clothing, and composition.
    \item \textbf{Select Weather:} Choose a specific weather condition from the library below.
    \item \textbf{Generate:} Create a prompt that combines the original scene with the new weather atmosphere.
\end{enumerate}

\textbf{Weather Library:}
\begin{itemize}
    \item \textbf{Clear Sunny:} Clear blue sky, harsh bright sunlight.
    \item \textbf{Heavy Rain:} Dark grey sky, pouring rain, splashing water, puddles.
    \item \textbf{Cloudy:} Flat grey clouds covering the sky, soft diffused light.
    \item \textbf{Snowy:} Falling snowflakes, ground covered in white snow.
    \item \textbf{Sunny Rain:} Bright sunshine shining while it is raining.
    \item \textbf{Foggy:} Thick mist obscuring the background, low visibility.
\end{itemize}

\textbf{Prompt Formula:} \\
{[Description of Subject and Scene] + [New Weather Details] + [Structure Preservation Constraint]}

\textbf{Preservation Rules:} \\
Use natural phrases to keep the original content:
\begin{itemize}
    \item "...while strictly maintaining the exact same composition, pose, and layout."
    \item "...keeping the original subject and camera angle unchanged."
\end{itemize}

\textbf{Output Format:}
\begin{itemize}
    \item \textbf{Target Weather:} [Selected Weather]
    \item \textbf{Prompt:} [The Final English Prompt]
\end{itemize}
\end{promptbox}

We utilized the generated prompts and the original image to generate the edited image. Then, we stitched these images together to create a video, using the image as the first frame. The prompt is as follows:

\begin{promptbox}{Image-to-Video Generation Prompt}
Cinematic motion, bring the image to life naturally. Subtle and fluid movement, maintaining strict fidelity to the input image. High temporal consistency, seamless visual flow, no morphing, no distortion. Consistent lighting and exposure, preserved subject identity, smooth camera motion, natural physics, continuous action.
\end{promptbox}

\section{Category Framework}

The detailed category framework is presented in Tables~\ref{tab:visual_attributes} and~\ref{tab:technical_attributes}. Table~\ref{tab:visual_attributes} covers \textbf{Subject}, \textbf{Style}, and \textbf{Background} attributes. Table~\ref{tab:technical_attributes} presents \textbf{Motion}, \textbf{Position}, \textbf{Camera}, and \textbf{Playback Technique} attributes.

% --- Table 1: Visual Content ---
\begin{table*}[htbp]
\centering
\small
\renewcommand{\arraystretch}{1.18}
\caption{Fine-grained subcategories for Subject, and Style.}
\label{tab:visual_attributes}

% --- 1. Subject ---
\begin{tabular}{>{\raggedright\bfseries}p{0.18\textwidth} >{\raggedright\arraybackslash}p{0.79\textwidth}}
\toprule
\multicolumn{2}{c}{\textbf{Subject}} \\ 
\midrule
Category & Description \\
\midrule
Type & Subject classification: persons, animals, plants, vehicles, buildings, virtual characters. \\
Count & Total number of subjects and per-category counts. \\
Appearance & Physical attributes: age, gender, ethnicity, physique, facial features, hairstyle, makeup. \\
Clothing & Attire and accessories: hats, glasses, jewelry, masks. \\
Pose & Body posture (standing, sitting), hand gestures, facial expressions. \\
State & Physical condition of inanimate objects (open/closed, broken). \\
Material & Material composition and texture: metal, wood, fabric, plastic. \\
Color & Dominant colors of objects, clothing, or skin tones. \\
OCR & Visible text transcription: signage, subtitles, logos, documents. \\
\bottomrule
\end{tabular}

\vspace{0.28cm}

% --- 2. Style ---
\begin{tabular}{>{\raggedright\bfseries}p{0.18\textwidth} >{\raggedright\arraybackslash}p{0.79\textwidth}}
\toprule
\multicolumn{2}{c}{\textbf{Style}} \\ 
\midrule
Category & Description \\
\midrule
Restricted Style Descriptors & Each video is categorized into one predefined style: \newline
\textbf{Traditional \& Fine Art:} Oil Painting, Watercolor, Sketch, Ink Wash, Ukiyo-e, Graffiti. \newline
\textbf{Digital \& Graphical:} CG Rendering, Pixel Art, Voxel Art, Low Poly, Minimalist, Flat Design. \newline
\textbf{Pop Culture \& Thematic:} Anime, Ghibli, American Comic, Cyberpunk, Steampunk, Retro. \newline
\textbf{Visual Tone:} Black and White, Monochromatic, Realistic. \\
\bottomrule
\end{tabular}

\vspace{0.28cm}

% --- 3. Background ---

\end{table*}

% --- Table 2: Dynamics & Technical ---
\begin{table*}[htbp]
\centering
\small
\renewcommand{\arraystretch}{1.18}
\caption{Subcategories for Background, Motion, Position, Camera, and Playback Technique.}
\label{tab:technical_attributes}

\begin{tabular}{>{\raggedright\bfseries}p{0.18\textwidth} >{\raggedright\arraybackslash}p{0.79\textwidth}}
\toprule
\multicolumn{2}{c}{\textbf{Background}} \\ 
\midrule
Category & Description \\
\midrule
Location & Scene type: office, street, park, indoor/outdoor environment. \\
Atmosphere & Mood or ambiance: festive, quiet, tense. \\
Lighting & Light source, time of day, style, brightness level. \\
Weather & Environmental conditions: sunny, rainy, snowy, foggy. \\
Key Background Objects & Prominent objects forming the main visual setting. \\
\bottomrule
\end{tabular}

% --- 4. Motion ---
\begin{tabular}{>{\raggedright\bfseries}p{0.18\textwidth} >{\raggedright\arraybackslash}p{0.79\textwidth}}
\toprule
\multicolumn{2}{c}{\textbf{Motion}} \\ 
\midrule
Category & Description \\
\midrule
Type & Primary movement or action sequence. \\
Interaction & Relationships between subjects. \\
Direction & Spatial orientation: moving left, approaching, receding. \\
Speed & Velocity: static, slow, fast, accelerating. \\
Amplitude & Magnitude, range, or force of motion. \\
Trajectory & Path of motion: linear, curved, circular. \\
\bottomrule
\end{tabular}

\vspace{0.28cm}

% --- 5. Position ---
\begin{tabular}{>{\raggedright\bfseries}p{0.18\textwidth} >{\raggedright\arraybackslash}p{0.79\textwidth}}
\toprule
\multicolumn{2}{c}{\textbf{Position}} \\ 
\midrule
Category & Description \\
\midrule
Subject-Object Relation & Spatial arrangement between subject and background elements. \\
Subject-Subject Relation & Relative positions and distances between multiple subjects. \\
Spatial Flipping & Geometric transformations: horizontal or vertical mirroring. \\
\bottomrule
\end{tabular}

\vspace{0.28cm}

% --- 6. Camera ---
\begin{tabular}{>{\raggedright\bfseries}p{0.18\textwidth} >{\raggedright\arraybackslash}p{0.79\textwidth}}
\toprule
\multicolumn{2}{c}{\textbf{Camera}} \\ 
\midrule
Category & Description \\
\midrule
Scale & Subject's frame size: close-up, medium, wide shot. \\
Movement & Camera motion: pan, tilt, zoom, dolly. \\
Orientation & Subject's direction relative to camera: front-facing, side profile. \\
Angle & Vertical camera position: eye-level, high angle, overhead. \\
Composition & Subject placement within frame: centered, left/right-aligned. \\
Depth of Field & Background clarity: blurred (shallow), sharp (deep). \\
Perspective & Point of view: first-person, third-person. \\
Shot Count & Single continuous shot vs. multi-shot sequence. \\
\bottomrule
\end{tabular}

\vspace{0.28cm}

% --- 7. Playback Technique ---
\begin{tabular}{>{\raggedright\bfseries}p{0.18\textwidth} >{\raggedright\arraybackslash}p{0.79\textwidth}}
\toprule
\multicolumn{2}{c}{\textbf{Playback Technique}} \\ 
\midrule
Category & Description \\
\midrule
Slow Motion & Reduced playback speed to emphasize details. \\
Fast Motion & Increased playback speed to compress timeline. \\
Reverse & Backward playback from end to start. \\
\bottomrule
\end{tabular}

\end{table*}

\end{document}